\documentclass[10pt,twocolumn,letterpaper]{article}

\usepackage[pagenumbers]{cvpr} 

\makeatletter
\@namedef{ver@everyshi.sty}{}
\makeatother
\usepackage{tikz}
\usepackage{graphicx}
\usepackage{amsmath}
\usepackage{amssymb}
\usepackage{booktabs}
\usepackage{multirow}
\usepackage{xspace}
\usepackage{nicematrix}
\usepackage[accsupp]{axessibility}

\usepackage[pagebackref,breaklinks,colorlinks]{hyperref}

\usepackage[capitalize]{cleveref}

\usepackage{sg-macros}
\newcommand{\dataset}{IAW\xspace}

\begin{document}

\title{Aligning Step-by-Step Instructional Diagrams to Video Demonstrations}

\author{
Jiahao Zhang$^{1,}$\thanks{Supported by an ANU-MERL PhD scholarship agreement.}\quad
Anoop Cherian$^{2}$\quad
Yanbin Liu$^{1}$\\
Yizhak Ben-Shabat$^{1,3,}$\thanks{Supported by Marie Sklodowska-Curie grant agreement No.~893465.}\quad
Cristian Rodriguez$^{4}$\quad
Stephen Gould$^{1,}$\thanks{Supported by an ARC Future Fellowship No.~FT200100421.}\\
$^1$The Australian National University,
$^2$Mitsubishi Electric Research Labs\\
$^3$Technion Israel Institute of Technology,
$^4$The Australian Institute for Machine Learning\\
{\tt\small $^1$\{first.last\}@anu.edu.au}
{\tt\small $^2$cherian@merl.com}
{\tt\small $^3$sitzikbs@gmail.com}
{\tt\small $^4$crodriguezop@gmail.com}\\
{\tt\small \href{https://davidzhang73.github.io/en/publication/zhang-cvpr-2023/}{https://davidzhang73.github.io/en/publication/zhang-cvpr-2023/}}
}

\maketitle

\begin{abstract}
    Multimodal alignment facilitates the retrieval of instances from one modality when queried using another. In this paper, we consider a novel setting where such an alignment is between (i) instruction steps that are depicted as assembly diagrams (commonly seen in Ikea assembly manuals) and (ii) segments from in-the-wild videos; these videos comprising an enactment of the assembly actions in the real world. We introduce a supervised contrastive learning approach that learns to align videos with the subtle details of assembly diagrams, guided by a set of novel losses. To study this problem and evaluate the effectiveness of our method, we introduce a new dataset: \dataset---for Ikea assembly in the wild---consisting of 183 hours of videos from diverse furniture assembly collections and nearly 8,300 illustrations from their associated instruction manuals and annotated for their ground truth alignments. We define two tasks on this dataset: First, nearest neighbor retrieval between video segments and illustrations, and, second, alignment of instruction steps and the segments for each video. Extensive experiments on \dataset demonstrate superior performance of our approach against alternatives.
\end{abstract}

\section{Introduction}
\label{sec:introduction}

The rise of \emph{Do-It-Yourself} (DIY) videos on the web has made it possible even for an unskilled person (or a skilled robot) to imitate and follow instructions to complete complex real world tasks~\cite{bonardi2020learning,liu2018imitation,torabi2019recent}.
One such task that is often cumbersome to infer from instruction descriptions yet easy to imitate from a video is the task of assembling furniture from its parts.
Often times the instruction steps involved in such a task are depicted in pictorial form, so that they are comprehensible beyond the boundaries of language (e.g.,~Ikea assembly manuals).
However, such instructional diagrams can sometimes be ambiguous, unclear, or may not match the furniture parts at disposal due to product variability.
Having access to video sequences that demonstrate the precise assembly process could be very useful in such cases.

Unfortunately, most DIY videos on the web are created by amateurs and often involve content that is not necessarily related to the task at hand.
For example, such videos may include commentary about the furniture being assembled, or personal assembly preferences that are not captured in the instruction manual.
Further, there could be large collections of videos on the web that demonstrate the assembly process for the same furniture but in diverse assembly settings; watching them could consume significant time from the assembly process.
Thus, it is important to have a mechanism that can effectively align relevant video segments against the instructions steps illustrated in a manual.

\begin{figure}[t]
    \centering
    \includegraphics[width=\linewidth]{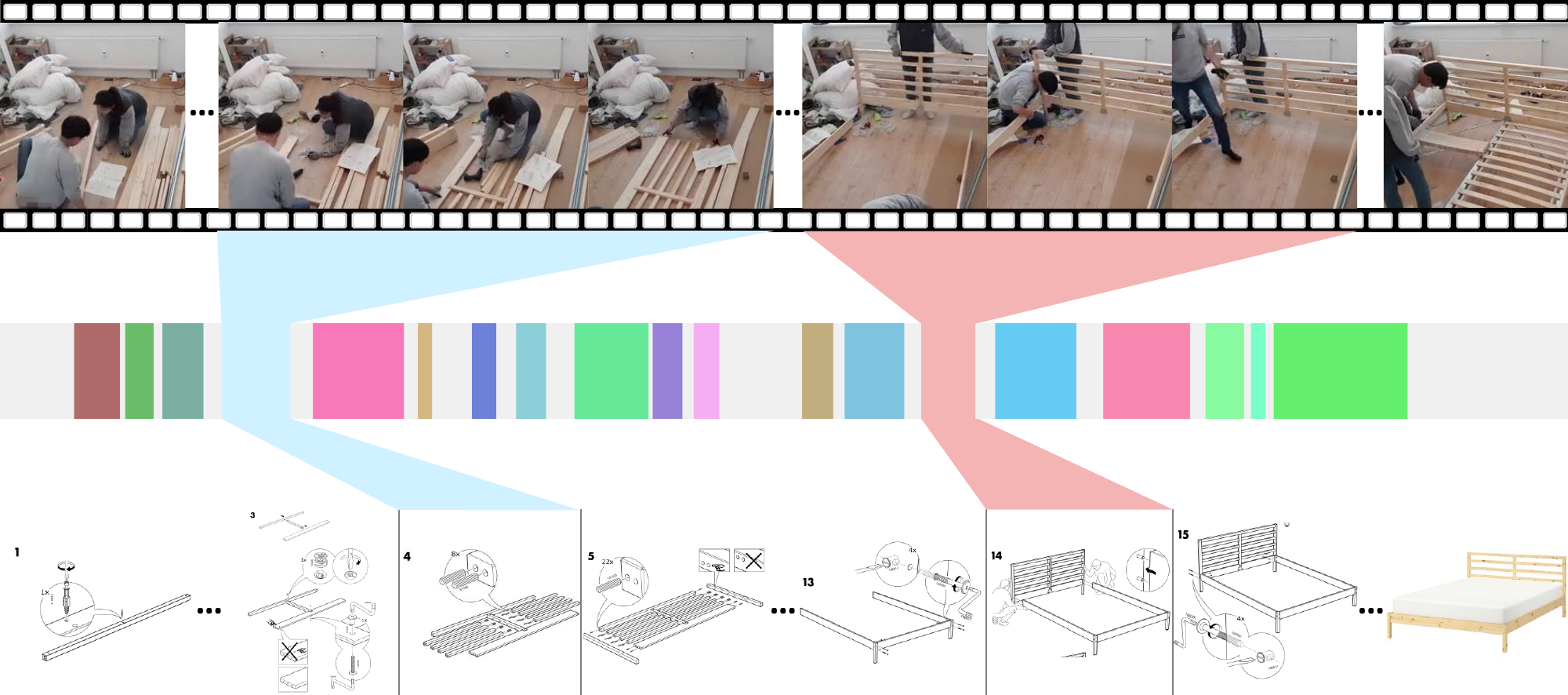}
    \caption{An illustration of video-diagram alignment between a YouTube video (top) \href{https://www.youtube.com/watch?v=He0pCeCTJQM}{He0pCeCTJQM} and an Ikea furniture manual (bottom) \href{https://www.ikea.com/au/en/p/tarva-bed-frame-pine-luroey-s49069795/}{s49069795}.}
    \label{fig:demo}
\end{figure}

In this paper, we consider this novel task as a multimodal alignment problem~\cite{radford2021learning,luo2022clip4clip}, specifically for aligning in-the-wild web videos of furniture assembly and the respective diagrams in the instruction manuals as shown in~\figref{fig:demo}.
In contrast to prior approaches for such multimodal alignment, which usually uses audio, visual, and language modalities, our task of aligning images with video sequences brings in several unique challenges.
First, instructional diagrams can be significantly more abstract compared to text and audio descriptions.
Second, illustrations of the assembly process can vary subtly from step-to-step (e.g., a rectangle placed on another rectangle could mean placing a furniture part on top of another).
Third, the assembly actions, while depicted in a form that is easy for humans to understand, can be incomprehensible for a machine.
And last, there need not be common standard or visual language followed when creating such manuals (e.g., a furniture piece could be represented as a rectangle based on its aspect ratio, or could be marked with an identifier, such as a part number).
These issues make automated reasoning of instruction manuals against their video enactments extremely challenging.

In order to tackle the above challenges, we propose a novel contrastive learning framework for aligning videos and instructional diagrams, which better suits the specifics of our task.
We utilize two important priors---a video only needs to align with its own manual and adjacent steps in a manual share common semantics---that we encode as terms in our loss function with multimodal features computed from video and image encoder networks.

To study the task in a realistic setting, we introduce a new dataset as part of this paper, dubbed \dataset for Ikea assembly in the wild.
Our dataset consists of nearly 8,300 illustrative diagrams from 420 unique furniture types scraped from the web and 1,005 videos capturing real-world furniture assembly in a variety of settings.
We used the Amazon Mechanical Turk to obtain ground truth alignments of the videos to their instruction manuals.
The videos involve significant camera motions, diverse viewpoints, changes in lighting conditions, human poses, assembly actions, and tool use.
Such in-the-wild videos offer a compelling setting for studying our alignment task within its full generality and brings with it a novel research direction for exploring the multimodal alignment problem with exciting real-world applications, e.g., robotic imitation learning, guiding human assembly, etc.

To evaluate the performance of our learned alignment, we propose two tasks on our dataset: (i) nearest neighbor retrieval between videos and instructional diagrams, and (ii) alignment of the set of instruction steps from the manual to clips from an associated video sequence.
Our experimental results show that our proposed approach leads to promising results against a compelling alternative, CLIP~\cite{radford2021learning}, demonstrating 9.68\% improvement on the retrieval task and 12\% improvement on the video-to-diagram alignment task.

\section{Related Work}
\label{sec:background}

\begin{figure*}
    \centering
    \begin{tabular}{cc}
        \includegraphics[height=5cm]{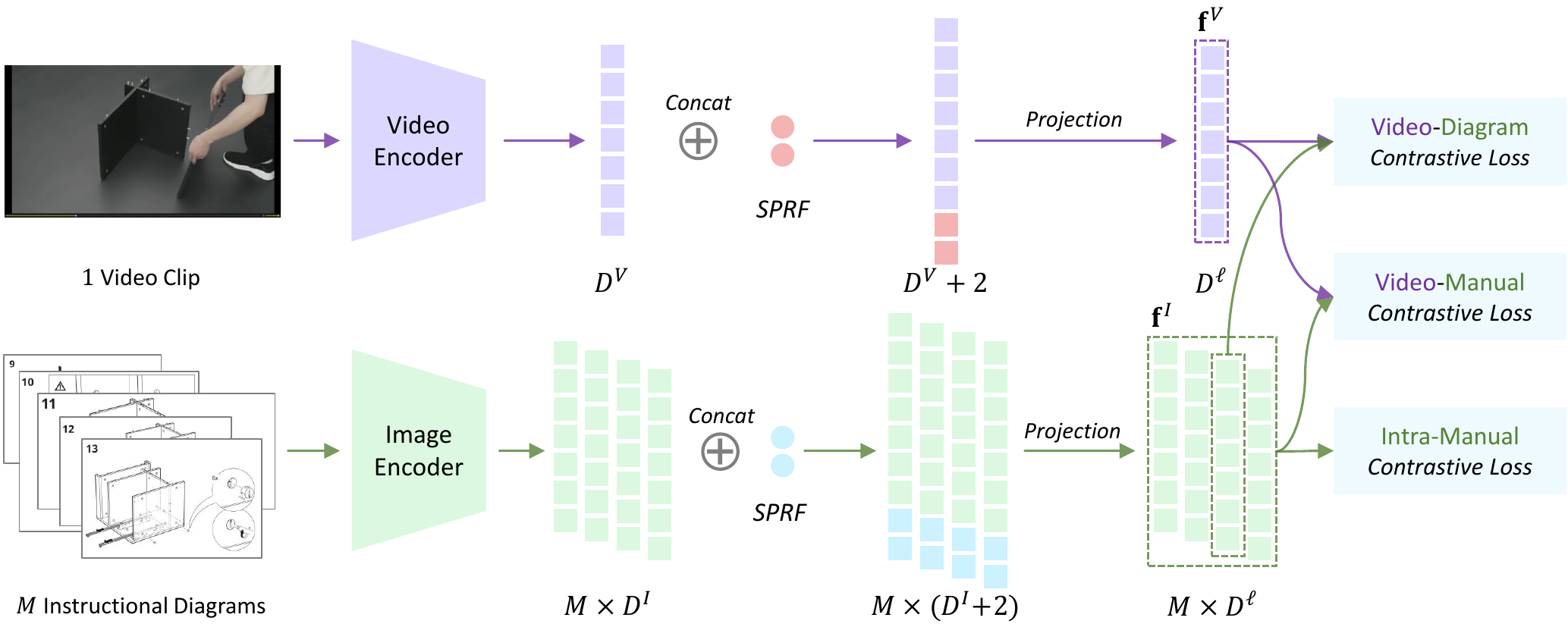} &
        \includegraphics[height=5cm]{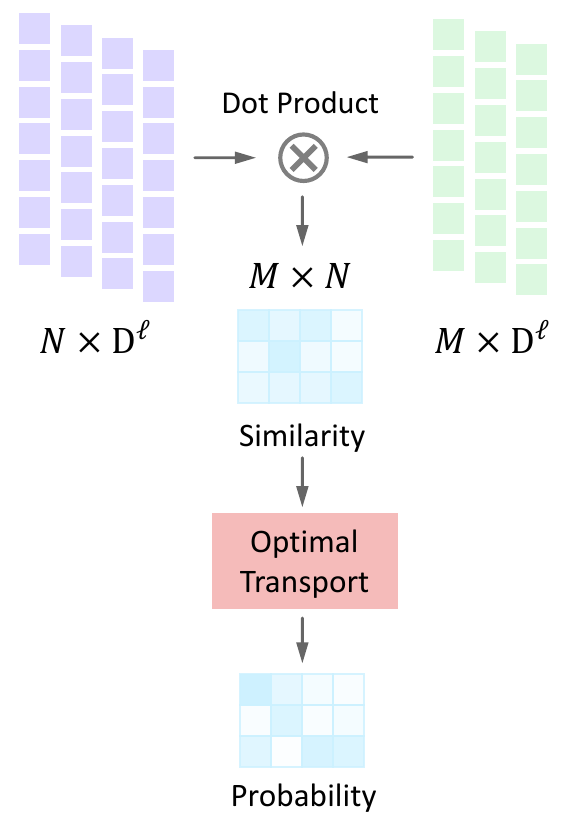}
        \\ {\footnotesize (a) Training stage.} & {\footnotesize{(b) Inference stage.}}
    \end{tabular}
    \caption{Schematic of our overall architecture.
        During training we extract features from each input video clip and set of instructional diagrams, respectively, using pre-trained encoders. We concatenate these with sinusoidal progress rate features (SPRF) introduced in~\secref{sec:sprf} and project into the same $D^\ell$ dimensionality space. The matched video clip and instructional diagram feature pairs are used for Video-Diagram Contrastive Loss, the video clip feature and $M$ instructional diagram features are fed into Video-Manual Contrastive Loss, and $M$ instructional diagram features themselves are used in Intra-Manual Contrastive Loss introduced in~\secref{sec:losses}. During inference all video features from $N$ sequential video clips, and all $M$ instructional diagram features from the corresponding manual are computed. We then form a similarity matrix and apply optimal transport (OT) introduced in~\secref{sec:ot} to produce the final alignment probabilities.}
    \label{fig:model}
\end{figure*}

\noindent\textbf{Assembly and Instructional Datasets.}
Multimodal video datasets (e.g., \cite{zhou2018towards, damen2018scaling, toyer2017human, ben2021ikea, wang2022translating, shao2016dynamic}) bridge the gap between video and other modalities such as the narratives from the video or instruction texts.
Among them, EPIC Kitchens~\cite{damen2018scaling} and YouCook2~\cite{zhou2018towards} align each video clip with the cooking procedure narratives.
Our dataset is more closely related to IKEA ASM~\cite{ben2021ikea} and IKEA-FA~\cite{toyer2017human}, which demonstrate furniture assembly instructions.
There are some other datasets focusing on converting assembly manuals to more comprehensible formats.
LEGO~\cite{wang2022translating} demonstrates how to obtain an executable plan from the assembly manuals while Shao et al.~\cite{shao2016dynamic} parses furniture assembly instructions into 3D models based on their manuals.
Unlike all of the above datasets, the proposed \dataset dataset aims to achieve the novel multimodal task of aligning in-the-wild web videos with step-by-step instructional diagrams.

\noindent\textbf{Multimodal Alignment.}
The classic work of Everingham et al.~\cite{everingham2006hello} focuses on aligning subtitle-transcript with person IDs in videos. Later works~\cite{tapaswi2015book2movie, dogan2018neural} started aligning video segments with text story-lines.
Recently, different approaches have been proposed for the text-video retrieval task, e.g., extracting fine-grained text features~\cite{wu2021hanet, han2021fine}, augmenting with more modalities~\cite{gabeur2020multi, wang2021t2vlad}, and contrastive text-video learning~\cite{liu2021hit, luo2022clip4clip, cheng2021improving, bogolin2022cross}.
Among them, Han et al.~\cite{han2022temporal} tackles alignment between assembly videos and text manuals.
However, due to the modality distinctions between text and image, these methods cannot be directly adopted to solve our problem.
Apart from the video modality, sketch images are similar to our instructional diagrams in the sense that both are black-white, text-free, and highly iconic abstract images.
Sketch-based video retrieval~\cite{collomosse2009storyboard} aims to retrieve specific video clips given a sketch image or sequence.
A recent related work to ours is Xu et al.~\cite{xu2020fine}, which extracts image features from both sketches and motion vector images, and optical flow from video clips.
They apply a triplet loss and a relation module on these extracted multimodal features to train the model.
However, their motion vector sketch is ad-hoc to specific video types, such as sports.
Compared with Xu et al.~\cite{xu2020fine}, our method is more general and supports two tasks from both video-to-diagram and diagram-to-video directions.

\noindent\textbf{Contrastive Learning.}
Contrastive learning was first introduced for self-supervised representation learning~\cite{oord2018representation, chen2020simple, he2020momentum, grill2020bootstrap, chen2021exploring}.
Then, the idea was naturally adapted for multimodal learning tasks, such as text-image alignment~\cite{radford2021learning} and text-video retrieval~\cite{luo2022clip4clip}.
CLIP~\cite{radford2021learning} designs a contrastive pre-training approach by predicting the correct pairs between images and their captions.
Since CLIP verifies the effectiveness of cross-modality contrastive learning, recent works~\cite{xue2022clip, bain2021frozen, wang2022omnivl} have incorporated it into the related models, facilitating the cross-modality video retrieval task.
Different from existing works, our alignment problem not only requires good contrastive between video clips and instructional diagrams but also entails distinguishing the subtle details in step-by-step manuals.
This motivates us to design three task-specific contrastive losses.

\section{Video-Instruction Alignment}
\label{sec:method}

\begin{figure*}
    \centering
    \includegraphics[width=0.99\linewidth]{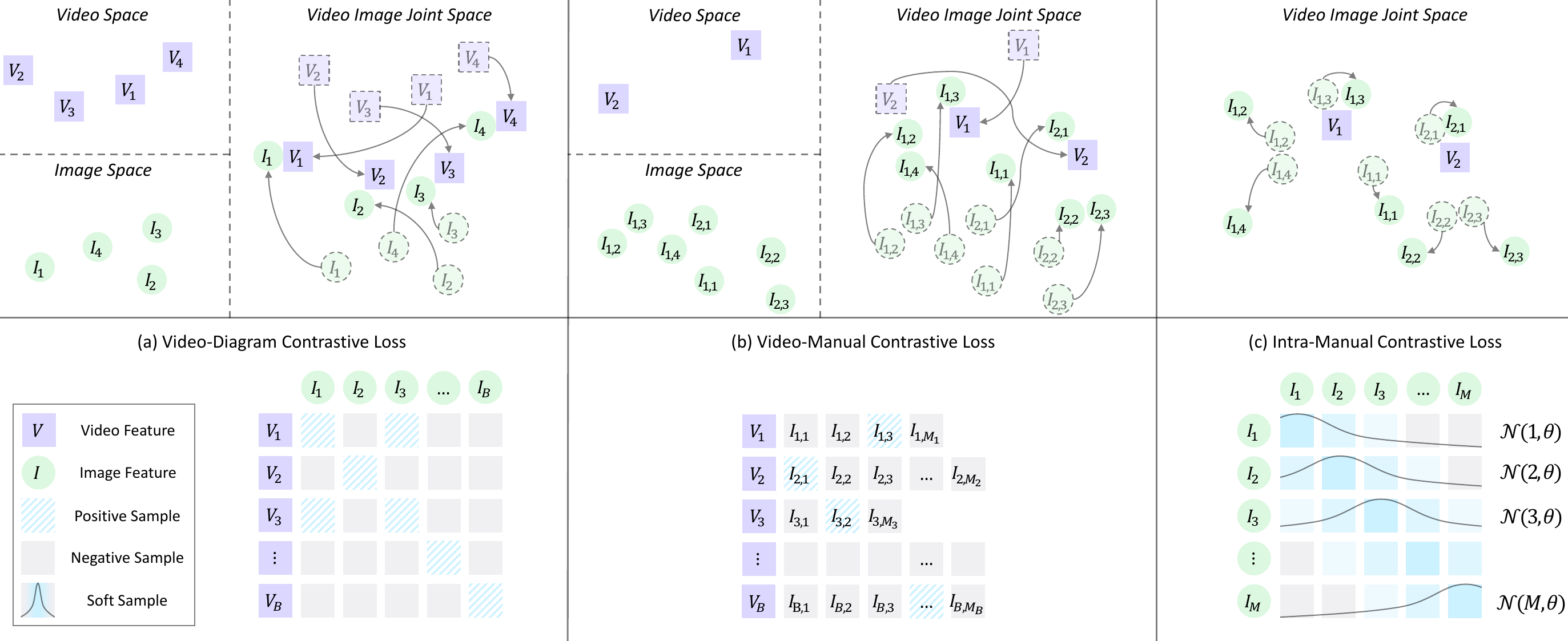}
    \caption{Visualization of our three losses described in~\secref{sec:losses}. The intent is depicted in the first row and the batch formation in the second row. Loss (a) tries to pair video and image up across the entire dataset. Loss (b) only matches video clips and images corresponding to the same manual. And loss (c) push images from the same manual apart from each other for better feature discrimination.}
    \label{fig:losses}
\end{figure*}

We formulate the task of aligning video segments to instructional diagrams as a variant of video-to-image matching.
Here the idea is to retrieve the image from a candidate set that most closely depicts the activity occurring in the short video clips and vice versa.
Importantly, since different instructional videos will involve different numbers of steps the candidate set will necessarily have variable cardinality (unlike, say, multi-class classification tasks).

Formally, given a set of $N$ video clips $\{V_i\}_{i=1}^{N}$ and a set of $M$ instructional diagrams $\{I_j\}_{j=1}^{M}$, our goal is to train a model to predict the correspondence between diagrams and clips.
A standard approach for addressing this problem is to learn a joint embedding space for videos and diagrams such that matching video-diagram pairs map near to each other in the embedding space.
Let $\mathbf{f}_i^\text{V}$ and $\mathbf{f}_j^{\text{I}}$ denote the feature embedding for the $i$-th video clip and $j$-th instructional diagram, respectively, and let $f_{\text{sim}}$ be some similarity measure.
Then, once the embedding space is learned we can use the model to predict the index of the instructional diagram corresponding to a given video clip $V$ as
\begin{align}
    j^\star & = \mathop{\text{argmax}}_{j=1,\dots,M} f_{\text{sim}}(\mathbf{f}^\text{V}, \mathbf{f}_j^\text{I}).
\end{align}
Likewise, we can find the video segment that most closely matches a given instructional diagram $I$ as
\begin{align}
    i^\star & = \mathop{\text{argmax}}_{i=1,\dots,N} f_{\text{sim}}(\mathbf{f}_i^\text{V}, \mathbf{f}^\text{I}).
\end{align}
This can be generalized to top-$k$ retrieval.
Last, we can enforce matching constraints, such as through optimal transport or dynamic time warping if order information is available, to jointly match all clips in a video to all steps in an instruction manual.
\figref{fig:model} depicts the overall model.

In this work, we use cosine similarity for $f_{\text{sim}}$.
The embedding vectors $\mathbf{f}_i^\text{V}$ and $\mathbf{f}_j^\text{I}$ are computed using video and image encoders trained under a contrastive loss and optionally augmented with temporal features such as we now describe.

\subsection{Sinusoidal Progress Rate Feature}
\label{sec:sprf}
Instruction manuals contain an ordered sequence of steps that is typically, although not always, followed during the assembly process.
However, the time needed to perform each step varies greatly depending on complexity of the step and experience of the assembler.
This suggests a weak correlation between (proportional) timestamps in the video and progress through the assembly process.
We can make use of this prior by including temporal ordering information in the video and diagram feature representations.

Given a video clip $V$ sampled from a full video of length $t_\text{duration}$ seconds, with start time $t_\text{start}$ and end time $t_\text{end}$, we define the video progress rate $r^\text{V}$ of that video clip as
\begin{align}
    r^\text{V} & =(t_\text{start}+t_\text{end})/2 t_\text{duration}
\end{align}
and the instructional diagram progress rate $r^\text{I}$ for the $j$-th step from a manual with $M$ total steps is simply $ r^\text{I} = j / M$.
Because we are using a cosine similarity function $f_\text{sim}$, we map the progress feature onto a half circle so that high similarity score coincides with when they align.
The final sinusoidal progress rate feature (SPRF) is then
\begin{align}
    (\sin(\pi r^V), \cos(\pi r^V))
\end{align}
for video and similarly for the instructional diagram, which we append to the feature embeddings extracted from the respective encoders (see~\figref{fig:model}).
Before and after the concatenation, the features are L2 normalized to alleviate side-effect due to fluctuation of numerical value scale.
Two fully connected layers then project each modality feature into the same dimensional space to form representations $\mathbf{f}^I$ and $\mathbf{f}^V$ for further similarity comparison.

\subsection{Training Losses}
\label{sec:losses}

Starting with pre-trained video and image encoders we finetune our model using variants of contrastive learning, which has recently been made popular for cross-modal matching by models like CLIP~\cite{radford2021learning}.
In this setting mini-batches are constructed by sampling video clip-instructional diagram pairs $(V_i, I_i)$ to optimize an infoNCE loss~\cite{hadsell2006dimensionality,oord2018representation} where pairs $(V_i, I_i)$ are considered positive and $(V_i, I_j)$, $i \neq j$ are considered negative.
Here we sample randomly from all videos and instruction manuals in the training data.

Formally, for mini-batch containing $B$ pairs, define
\begin{align}
    p^{V2I}_{ij} & = \frac{\exp(f_\text{sim}(\boldf^V_i,\boldf^I_j)/\tau)}{\sum_{b=1}^B{\exp(f_\text{sim}(\boldf^V_i,\boldf^I_b)/\tau)}}
    \label{eqn:video_to_image}
    \\
    p^{I2V}_{ji} & = \frac{\exp(f_\text{sim}(\boldf^V_i,\boldf^I_j)/\tau)}{\sum_{b=1}^B{\exp(f_\text{sim}(\boldf^V_b,\boldf^I_j)/\tau)}}
    \label{eqn:image_to_video}
\end{align}
to be the probability of matching video $V_i$ to image $I_j$ and the probability of matching image $I_j$ to video $V_i$, respectively.
Here $\tau$ is a temperature parameter that controls the bias towards difficult examples~\cite{wang2021understanding}.
Standard contrastive learning then minimizes
\begin{align}
    \Ell_\text{infoNCE} & = -\frac{1}{2B}\left(\sum_{i=1}^B \log p^{V2I}_{ii} + \sum_{j=1}^B \log p^{I2V}_{jj}\right).
\end{align}

We note that this vanilla version of contrastive learning does not consider situations where there may be many-to-one matches between pairs.
Specifically, in our application multiple video clips may map to the same step.
We introduce a specialized loss to deal with this scenario.

\noindent\textbf{Video-Diagram Contrastive Loss (\figref{fig:losses}(a)).}
Contrastive learning frameworks benefit from large batch sizes~\cite{radford2021learning}.
However, as batch size increases there is a greater chance that we sample multiple videos matching to the same diagram within the batch, which violates the assumptions of the infoNCE loss.
To better handle these cases we build on the work of~\cite{wang2021actionclip} that introduces a Kullback-Leibler (KL) divergence loss between predicted and ground truth distributions, $\bp$ and $\bq$, respectively.
However, rather than KL-divergence, we prefer Jensen-Shannon (JS) divergence, which we find improves training stability.

Let $\bp^{V2I}$ and $\bp^{I2V}$ be vectors containing all video-to-diagram and diagram-to-video probabilities introduced in Eqs.
\ref{eqn:video_to_image} and \ref{eqn:image_to_video}, respectively.
Similarly, let $\bq^{V2I}$ and $\bq^{I2V}$ be the corresponding ground truth alignment distributions.
Then our video-diagram contrastive loss is defined as
\begin{align}
    \Ell^\text{VI} & = \frac{1}{2}\left(D_{JS}(\bp^{V2I} \| \bq^{V2I}) + D_{JS}(\bp^{I2V} \| \bq^{I2V})\right)
\end{align}
where $D_{JS}$ is the Jensen-Shannon divergence.

\noindent\textbf{Video-Manual Contrastive Loss (\figref{fig:losses}(b)).}
The above losses align video and diagram pairs globally across the entire training dataset.
However, for our task we know that a given video clip only needs to match against one of the steps in its corresponding instruction manual, not other manuals.
Hence, we can perform a more task-specific discrimination by exploiting this prior information in the model.
To do so we modify our procedure for constructing a mini-batch to first sample a video clip $V_i$ and then include all instructional diagrams $\{I_1, \ldots, I_{M_i}\}$ from the video's corresponding manual.
One of these diagrams will be the ground truth positive match for the clip.
We then employ a classification loss based on cross entropy (CE) as
\begin{align}
    \Ell^{VM} & = \sum_{i=1}^B\frac{M_i}{\sum_{b=1}^B M_b} CE(\bp_{i}^{V2I}, \bp_{i}^{gt})
\end{align}
where $M_i$ indicates the length of the manual corresponding to the $i$-th video.
Here $\bp_{i}^{V2I} \subseteq (p^{V2I}_{ij})_{j=1}^{B}$ is a subvector of probabilities for matching video $V_i$ to all diagrams $I_j$ from the corresponding manual and $\bp_i^{gt}$ is the associated one-hot ground truth encoding.
We weight each term in the loss by $\frac{M_i}{\sum_{b=1}^B M_b}$ to give more emphasis to more difficult assemblies, assumed to be the ones containing more steps.

\noindent\textbf{Intra-Manual Contrastive Loss (\figref{fig:losses}(c)).}
The previous losses only consider contrasting embeddings between videos and diagrams.
However, most furniture assembly tasks involve a progressive process where the visual similarity between successive steps is large.
Indeed, the main component of the assembly is often introduced early in the assembly process and dominates the instructional diagram.
This makes it challenging to distinguish between steps.
To encourage diagrams from the same manual to be spread out in embedding space, so that they are more easily distinguished, we introduce an intra-manual contrastive loss.

Similar to the video-to-diagram and diagram-to-video matching probabilities defined above, let
\begin{align}
    p^{I2I}_{jk} & = \frac{\exp(f_\text{sim}(\boldf^I_j,\boldf^I_k)/\tau)}{\sum_{m=1}^M{\exp(f_\text{sim}(\boldf^I_j,\boldf^I_m)/\tau)}}
\end{align}
be the probability of matching diagram $I_j$ to diagram $I_k$ from the same manual according to our similarity metric.
Then we define our intra-manual contrastive loss as
\begin{align}
    \Ell^M & = \sum_{j=1}^B\frac{M_j}{\sum_{b=1}^B M_b} D_{JS}\left(\bp_j^{I2I} \| \N(j, \theta)\right)
\end{align}
where $\bp_j^{I2I}$ is the softmax normalized diagram-to-diagram probability vector associated with diagram $I_j$,
and $\N(j, \theta)$ is a univariate Gaussian distribution with mean $j$, learnable variance $\theta$
and discretized on support $\{1, \ldots, M_j\}$.
This encourages distances in diagram embedding space to correspond to distances between steps in the manual.
We use a normal distribution instead of a delta distribution as a relaxation since nearby negative diagrams are still likely to share some semantics.

\subsection{Set Matching}
\label{sec:ot}
Our model is very general.
Given a single video clip we can retrieve the most likely diagram showing the assembly step and given a single diagram we can retrieve a set of best matching video clips.
To align an entire video (sequence of clips) to an entire instruction manual, we can add approximate one-to-one matching priors or temporal constraints, through optimal transport (OT)
or dynamic time warping (DTW), respectively.
As we will see in our experiments, the absence of temporal order constraints in OT slightly outperforms DTW due to occasional out-of-order execution of assembly steps or strong false matches.

To apply either method we first extract features $\boldf^V_i$ for an entire video $\{V_i\}_{i=1}^N$ and $\boldf^I_j$ for all instructional diagrams in the corresponding manual $\{I_j\}_{j=1}^M$.
Denote by $s_{ij}$ the similarity $f_\text{sim}(\boldf_i^V, \boldf_j^I)$ between video clip $V_i$ and diagram $I_j$.
Let $\overline{s} = \max_{i,j} s_{ij}$ and $\underline{s} = \min_{ij} s_{ij}$.
We then construct a cost matrix $C \in \reals^{N \times M}$ with entries
\begin{align}
    C_{ij} & = \frac{s_{ij}^\alpha - \underline{s}^\alpha}{\overline{s}^\alpha -  \underline{s}^\alpha}.
\end{align}
Here $\alpha > 1$ accentuates the similarity differences and the normalization by $\overline{s}^\alpha - \underline{s}^\alpha$ restricts the range of $C_{ij}$ to $[0, 1]$.
The optimal transportation plan $T^{\star}$ obtained by solving the entropy regularized optimal transport problem,
\begin{align}
    \begin{array}{rl}
        \text{minimize}   & \sum_{i=1}^{N} \sum_{j=1}^{M} T_{ij} C_{ij} - \epsilon H(T)           \\
        \text{subject to} & \sum_{i=1}^{M} T_{ij} = \frac{1}{N}, \,\text{for $j = 1, \ldots, N$}  \\
                          & \sum_{j=1}^{N} T_{ij} = \frac{1}{M}, \,\text{for $i = 1, \ldots, M$},
    \end{array}
\end{align}
gives the joint probability of matching videos and diagrams.
It can be found efficiently by applying the Sinkhorn-Knopp algorithm~\cite{sinkhorn1967diagonal} to the optimization problem defined above.

In a similar fashion, we can use DTW to find the optimal path through the cost matrix to give the most likely matching subject to the ordering constraint that later video clips cannot match to earlier instructional diagrams and vice versa.
More formally, if video clip $V_i$ matches to diagram $I_j$ then clip $V_{i+1}$ cannot match to diagram $I_{j'}$ with $j' < j$ and diagram $I_{j+1}$ cannot match to video clip $V_{i'}$ with $i' < i$.

\section{Ikea Assembly in the Wild Dataset (\dataset)}
\label{sec:dataset}

In order to study the problem of understanding instructional videos,
we collected a large well-labeled dataset called the Ikea assembly in-the-wild (\dataset) dataset with annotations obtained using Amazon Mechanical Turk and a publicly available in-browser video annotation tool Vidat~\cite{zhang2020vidat}.
The \dataset dataset contains 420 Ikea furniture pieces from 14 common categories, e.g., sofa, bed, wardrobe, table, etc.
Each piece of furniture comes with one or more user instruction manuals,
which are first divided into pages and then further divided into independent steps cropped from each page (some pages contain more than one step and some pages do not contain instructions).
There are 8,568 pages and 8,263 steps overall, on average 20.4 pages and 19.7 steps for each piece of furniture.
We crawled YouTube to find videos corresponding to these instruction manuals and as such the conditions in the videos are diverse on many aspects,
e.g.,~duration, resolution, first- or third-person view, camera pose, background environment, number of assemblers, etc.
The IAW dataset contains 1,005 raw videos with a length of around 183 hours in total.
Among them, approximately 114 hours of content are labeled as 15,649 actions to match the corresponding step in the corresponding manual.

The dataset is split into a train, validation, and test set (with 30,876 segments, 6,871 segments and 11,103 segments, respectively) by using a greedy algorithm to balance the distribution with respect to all attributes including viewpoint, indoor or not, camera motion and number of assemblers involved, and it is guaranteed that all video in both validation and testing sets are unseen in the training set.

\section{Experiments}
\label{sec:experiments}

\begin{table*}[t]
    \centering
    \small
    \newcolumntype{Z}{>{\centering\arraybackslash}X}
    \caption{Results comparing model alternatives. Performance on cropped \emph{step} diagrams is denoted by S and entire \emph{pages} from the manual by P. For fair comparison, the backbone for encoders is kept the same and only the loss and post-processing are varied. \textsc{CosSim} uses cosine similarity loss and CLIP uses infoNCE loss on paired features. \dag{} AUROC values below 0.5 come from the fact that not every step or page diagram has a corresponding video segment in the test set, i.e., some queries have no positives.}
    \resizebox{\linewidth}{!}{
        \begin{NiceTabularX}{\linewidth}{
                @{}
                w{l}{5em}
                *{4}{Z}
                |
                *{6}{Z}
            }
            \toprule
            \Block{3-1}{Method}
            & \Block{1-4}{Video to diagram retrieval}
            &
            &
            &
            & \Block{1-6}{Diagram to video retrieval}
            \\
            & \Block{1-2}{Top1 Acc.\%$\uparrow$}
            &
            & \Block{1-2}{AIE$\downarrow$}
            &
            & \Block{1-2}{R@1$\uparrow$}
            &
            & \Block{1-2}{R@3$\uparrow$}
            &
            & \Block{1-2}{AUROC$\uparrow$}
            \\
            \cmidrule(lr){2-3}\cmidrule(lr){4-5}\cmidrule(lr){6-7}\cmidrule(lr){8-9}\cmidrule(lr){10-11}
            & S
            & P
            & S
            & P
            & S
            & P
            & S
            & P
            & S
            & P
            \\
            \midrule
            Random
            & 5.664
            & 5.107
            & 9.334
            & 8.131
            & 6.576
            & 3.393
            & 19.90
            & 10.16
            & 0.375
            & 0.244
            \\
            \textsc{CosSim}
            & 11.89
            & 11.06
            & 4.360
            & 4.368
            & 12.43
            & 6.780
            & 32.90
            & 20.93
            & 0.561
            & 0.336
            \\
            CLIP
            & 19.61
            & 19.05
            & 4.274
            & 4.180
            & 16.94
            & 10.25
            & 38.67
            & 23.45
            & 0.590
            & 0.373
            \\\midrule
            Ours
            & 28.62
            & 34.55
            & 3.734
            & 2.928
            & 22.30
            & 16.48
            & 45.00
            & 32.20
            & 0.617
            & \hphantom{${}^\dag{}$}0.390${}^\dag{}$
            \\
            \quad w/o SPRF
            & 21.73
            & 27.08
            & 6.018
            & 4.485
            & 16.90
            & 13.17
            & 36.07
            & 26.70
            & 0.558
            & 0.357
            \\
            \quad w/ DTW
            & 31.45
            & 36.20
            & \textbf{3.382}
            & \textbf{2.752}
            & 23.20
            & 17.32
            & 32.45
            & 17.55
            & 0.467
            & 0.310
            \\
            \quad w/ OT
            & \textbf{31.61}
            & \textbf{36.71}
            & 3.458
            & 2.816
            & \textbf{26.62}
            & \textbf{18.28}
            & \textbf{49.11}
            & \textbf{32.28}
            & \textbf{0.626}
            & \textbf{0.401}
            \\\bottomrule
        \end{NiceTabularX}
    }
    \label{tbl:result}
\end{table*}

We evaluate our model on the \dataset dataset for tasks of finding the best instructional diagram for a given video clip (video-to-diagram retrieval) and finding the top-$k$ video clips corresponding to a given diagram (diagram-to-video retrieval).
We consider two settings: independent retrieval where we are given one query video (resp. diagram) at a time, and set retrieval where we are given an entire video and corresponding instruction manual.
The latter is the alignment problem and allows us to use structured inference methods such as optimal transport (OT) and dynamic time warping (DTW).

\begin{figure}[t]
    \centering
    \includegraphics[width=0.85\linewidth]{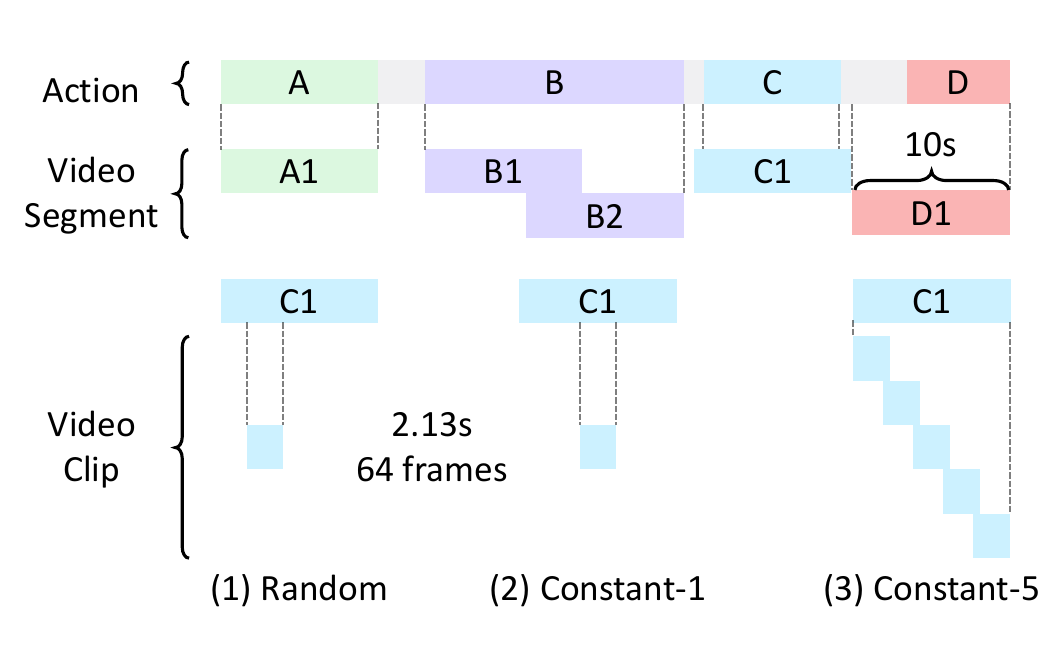}
    \caption{Demonstration of the video clip sampler for a task with four steps, A, B, C and D. Each step is sub-sampled to multiple 10s video segments (\eg B $\rightarrow$ B1 and B2) with back padding. As input to the video encoder, each video segment is further divided into 2.13s video clips. We randomly sample one clips for training (1), choose a constant single clip for validation (2), and average over five clips for testing (3).}
    \label{fig:sampler}
\end{figure}

\noindent\textbf{Preprocessing.} We re-sample all videos to 30fps and then sub-sample into 10s segments to align with common practice of action recognition tasks.
Video clips of duration 2.13s (64 frames) are used as input to the video encoder as shown in~\cref{fig:sampler}.
The short side of each video frame is down-sampled to 224, maintaining aspect ratio.
The long side of each instructional diagram is down-sampled to 224, also maintaining aspect ratio and padding the short side with white pixels.
Random resize crops are used as data augmentation for videos, and random resize crops, horizontal flips and rotations are applied to instructional diagrams.

\noindent\textbf{Architecture.} We choose ResNet-50~\cite{he2016deep} pretrained on ImageNet~\cite{deng2009imagenet} as our backbone image encoder, and a ResNet-50 based Kinetics 400~\cite{carreira2017quo} pretrained Slowfast-8x8~\cite{feichtenhofer2019slowfast} for video encoding.
For each 64-frame clip, 8 frames are uniformly sub-sampled for the slow path, and 32 frames for the fast path.
We remove the classification heads from these two backbone models and freeze the entire video encoder, but only first three layers of the image encoder allowing the later layers to be finetuned.
The dimensionality of both video and diagram features is set to 1024.

\noindent\textbf{Training Details.} A dedicated learnable temperature parameter $\tau$ is assigned to each loss and initialized to 0.07 following~\cite{wu2018unsupervised}.
The variance $\sigma$ in intra-manual contrastive loss is initialized to 1 to represent a standard normal distribution.
We use AdamW~\cite{loshchilov2017decoupled} as the optimizer with learning rate $5 \times 10^{-4}$ and weight decay $5 \times 10^{-3}$.
All models are trained for 20 epochs with 128 video clips per batch (and number of instructional diagram depending on the losses being used as described in~\secref{sec:method}).
We select the model from the epoch with highest top-1 accuracy on the validation set for reporting test set results.
It takes approximately 20 hours on a single Nvidia A100 GPU 80GB per experiment.
During alignment testing, similarity scores are aggregated into a single $N \times M$ matrix for each video and corresponding instruction manual and optimal transport applied with hyper-parameters $\epsilon=4$ and $\alpha=7$.

\noindent\textbf{Evaluation metrics.} We report average top-1 accuracy and average index error (AIE) for the video-to-diagram retrieval task on the test set.
AIE is useful for characterizing errors since predicting a step near to the ground truth is better than predicting one that is far away.
It is defined as,
\begin{align}
    \text{AIE} & = \frac{1}{N} \sum_{i=1}^N |j_i^\star - j_i^\text{gt}|
\end{align}
where $j_i^\star$ is the predicted diagram index for the $i$-th video and $j_i^\text{gt}$ is its true index.
Since a single instructional diagram can correspond to multiple video clips we adopt recall@1, recall@3 and area under the ROC curve as metrics for the diagram-to-video task.
Unless otherwise stated we report results at the video segment level.
Here we sample consecutive 64-frame video clips (2.13s) from each 10s video segment and average the features from the video encoder.

\subsection{Main Results}

Our main results are reported in~\tabref{tbl:result}.
We compare to two baseline methods, {\sc CosSim} and CLIP, which use a cosine similarity loss and infoNCE loss only on paired features.
We report results on four variants of our approach using all three losses described in~\secref{sec:method}.
The first (``Ours'') includes the sinusoidal progress rate features (SPRF) capturing temporal information.
Note that we also experimented with more standard positions encoding methods popular with transformers~\cite{vaswani2017attention} but found these to produce inferior results (omitted here for brevity).
The second variant (``w/o SPRF'') shows results with this feature removed.
The last two variants use dynamic time warping (DTW) and optimal transport (OT) in the alignment setting, i.e., with access to complete videos and instruction manuals.

\begin{figure}[t]
    \centering
    \includegraphics[width=\linewidth]{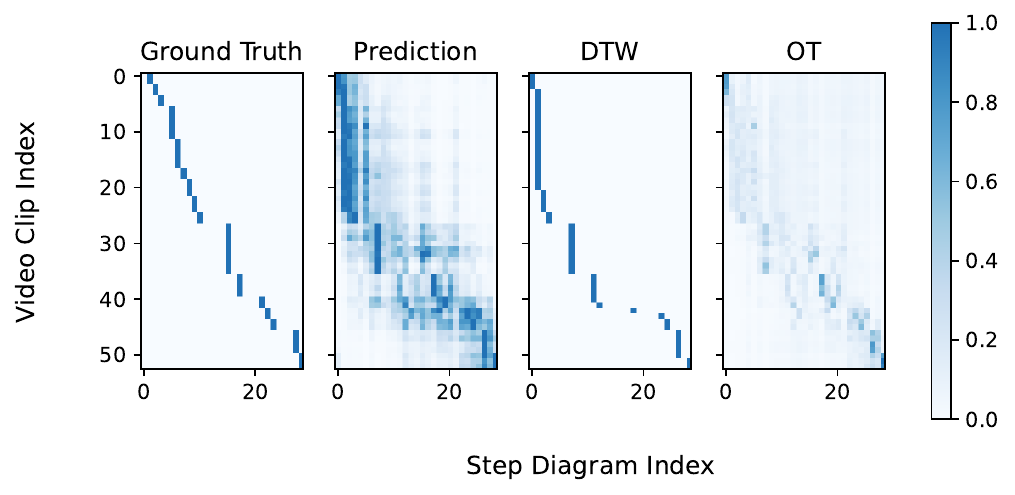}
    \caption{An example of post processing with DTW and OT with furniture \href{https://www.ikea.com/au/en/p/friheten-three-seat-sofa-bed-skiftebo-dark-grey-30341149/}{30341149} and video \href{https://www.youtube.com/watch?v=dzLNgz861Hk}{dzLNgz861Hk}.}
    \label{fig:ot}
\end{figure}

Observe that our method significantly outperforms the baseline approaches on both video-to-diagram retrieval and diagram-to-video retrieval.
This is largely due to our SPRF feature but also thanks to the improved loss functions (we provide a complete ablation analysis below).
Further improvement in performance can be gained by post processing with DTW or OT.
Interestingly, OT does slightly better than DTW indicating the the ordering constraint imposed by DTW is too restrictive for this task.
See~\figref{fig:ot} for an example alignment.

\begin{figure*}
    \centering
    \begin{tabular}{cc}
        \includegraphics[width=0.476\linewidth]{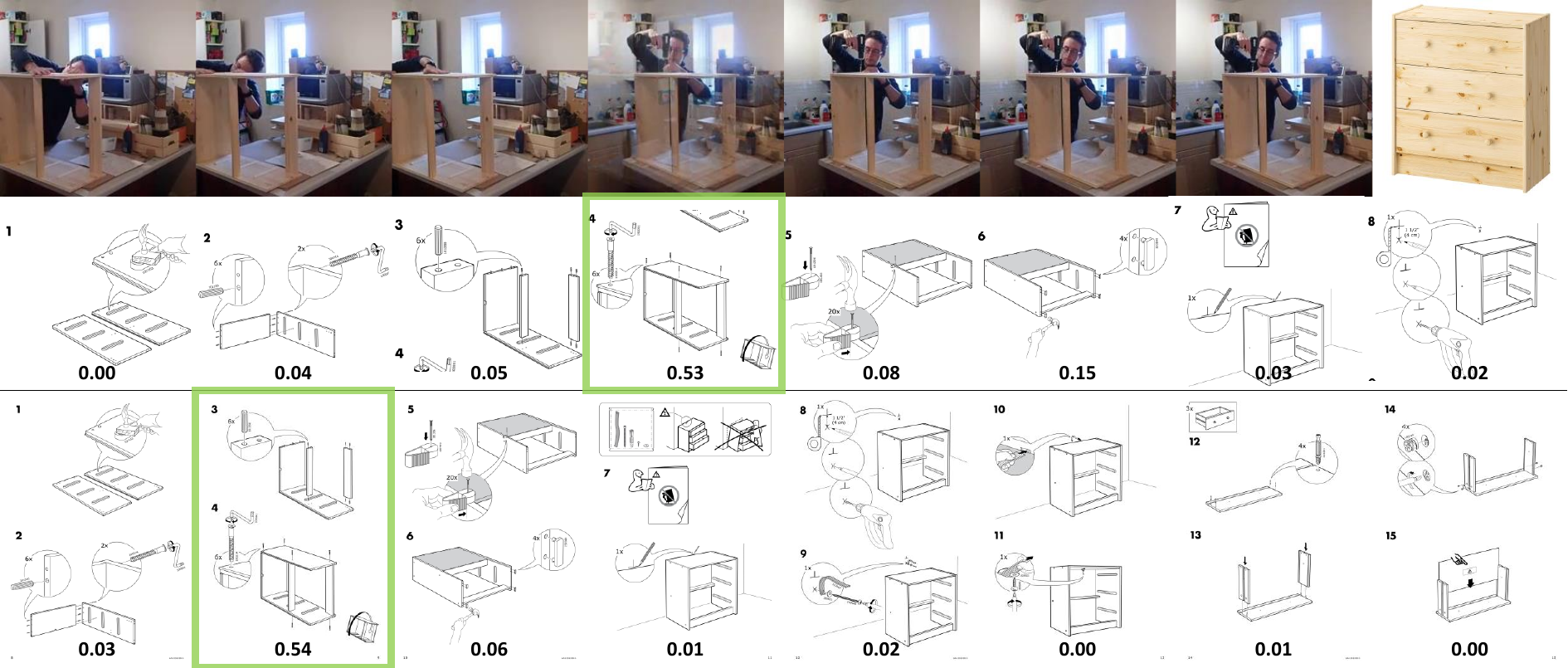}
         &
        \includegraphics[width=0.476\linewidth]{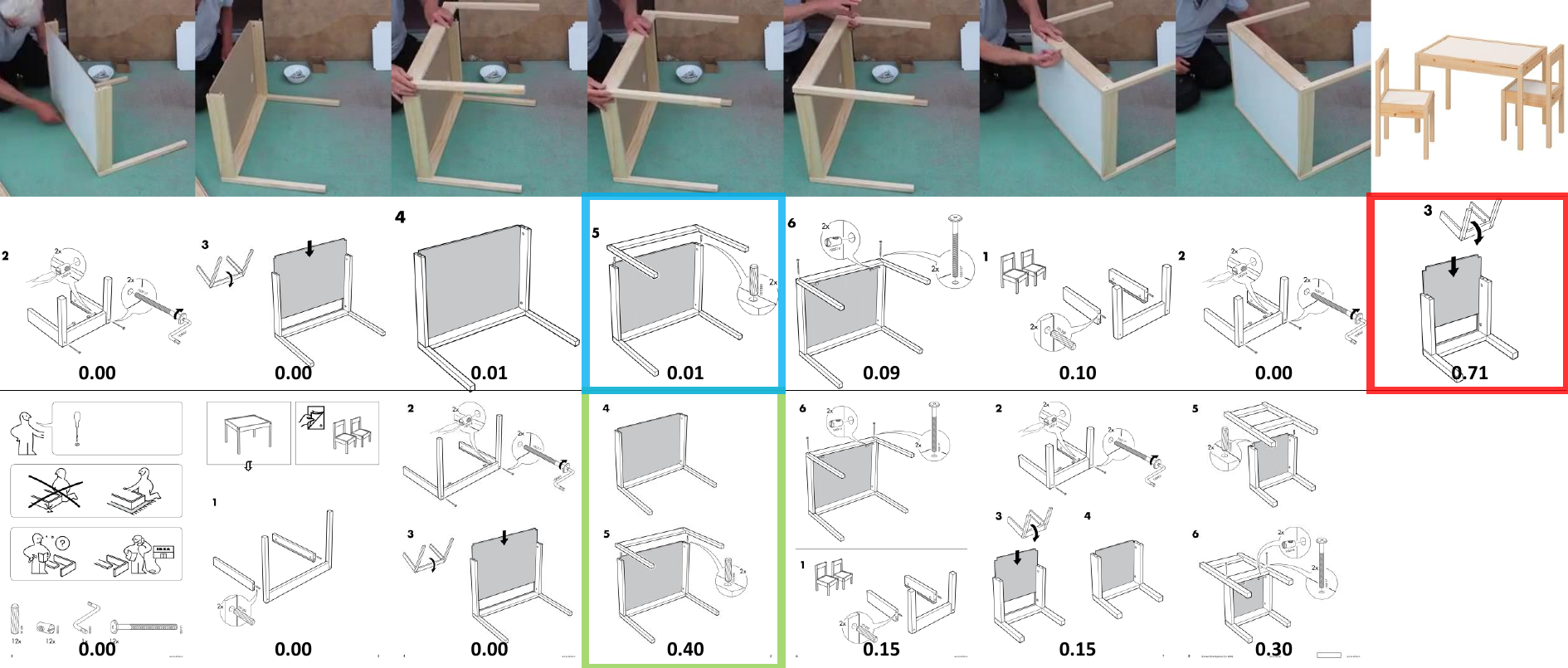}
        \\
        \parbox{0.476\linewidth}{\vspace{0.3cm} \small (a) Successful alignment between YouTube video \href{https://www.youtube.com/watch?v=moq_A1o3ZKw}{moq\_A1o3ZKw} and Ikea furniture manual \href{https://www.ikea.com/au/en/p/rast-chest-of-3-drawers-pine-60356219/}{60356219}.}
         &
        \parbox{0.476\linewidth}{\vspace{0.3cm} \small (b) A failed alignment between YouTube video \href{https://www.youtube.com/watch?v=d6sbVuHV0bc}{d6sbVuHV0bc} and Ikea furniture manual \href{https://www.ikea.com/au/en/p/laett-childrens-table-with-2-chairs-white-pine-10178413/}{10178413}.}
    \end{tabular}
    \caption{Qualitative results. Rows show frames from a single video clip; step instructional diagrams (subset shown); and page instructional diagrams. Prediction is highlighted by a green box for correct or a red box for incorrect; ground truth is then highlighted with a blue box.}
    \label{fig:qualitative}
\end{figure*}

Qualitative results are shown in~\figref{fig:qualitative}.
We show one correct alignment and one incorrect alignment for matching to assembly steps.
Notice the high degree of similarity between steps in the assembly process, which makes this an extremely challenging task.
Further examples are included on the project website for this paper.

\subsection{Effect of Losses}

\begin{table*}[t]
    \centering
    \small
    \newcolumntype{Z}{>{\centering\arraybackslash}X}
    \caption{Ablation analysis on different loss combinations reported without post processing. Batches have 128 video clips (plus diagrams as required) except for those marked with \dag{} where we double the number of video clips to 256 (plus a single matching diagram). See~\secref{sec:method}.}
    \resizebox{\linewidth}{!}{
        \begin{NiceTabularX}{\linewidth}{
                @{}
                w{l}{2.5em}
                *{6}{w{l}{0.4em}}
                *{4}{Z}
                |
                *{6}{Z}
            }
            \toprule
            \Block{3-1}{Exp.}
            &
            &
            &
            &
            &
            &
            & \Block{1-4}{Video to diagram retrieval}
            &
            &
            &
            & \Block{1-6}{Diagram to video retrieval}
            \\
            & \Block{1-2}{Loss A}
            &
            & \Block{1-2}{Loss B}
            &
            & \Block{1-2}{Loss C}
            &
            & \Block{1-2}{Top1 Acc.\%$\uparrow$}
            &
            & \Block{1-2}{AIE$\downarrow$}
            &
            & \Block{1-2}{R@1$\uparrow$}
            &
            & \Block{1-2}{R@3$\uparrow$}
            &
            & \Block{1-2}{AUROC$\uparrow$}
            \\
            \cmidrule(lr){2-3}\cmidrule(lr){4-5}\cmidrule(lr){6-7}\cmidrule(lr){8-9}\cmidrule(lr){10-11}\cmidrule(lr){12-13}\cmidrule(lr){14-15}\cmidrule(lr){16-17}
            & S
            & P
            & S
            & P
            & S
            & P
            & S
            & P
            & S
            & P
            & S
            & P
            & S
            & P
            & S
            & P
            \\\midrule
            \text{CosSim}\textsuperscript{\dag}
            &
            &
            &
            &
            &
            &
            & 11.89
            & 11.06
            & 4.360
            & 4.368
            & 12.43
            & 6.780
            & 32.90
            & 20.93
            & 0.561
            & 0.336
            \\
            CLIP\textsuperscript{\dag}
            &
            &
            &
            &
            &
            &
            & 19.61
            & 19.05
            & 4.274
            & 4.180
            & 16.94
            & 10.25
            & 38.67
            & 23.45
            & 0.590
            & 0.373
            \\\midrule
            A1\textsuperscript{\dag}
            & \checkmark
            &
            &
            &
            &
            &
            & 20.87
            & 15.28
            & 3.991
            & 4.635
            & 17.92
            & 8.650
            & 41.00
            & 22.17
            & 0.592
            & 0.358
            \\
            A2\textsuperscript{\dag}
            &
            & \checkmark
            &
            &
            &
            &
            & 19.02
            & 19.49
            & 4.086
            & 3.979
            & 17.42
            & 10.57
            & 38.99
            & 24.69
            & 0.577
            & 0.373
            \\
            A3\textsuperscript{\dag}
            & \checkmark
            & \checkmark
            &
            &
            &
            &
            & 20.58
            & 19.34
            & 4.036
            & 4.090
            & 17.08
            & 10.13
            & 39.89
            & 24.64
            & 0.583
            & 0.371
            \\\midrule
            B1
            &
            &
            & \checkmark
            &
            &
            &
            & 27.20
            & 20.74
            & 3.842
            & 4.160
            & 20.93
            & 10.52
            & 44.35
            & 25.29
            & 0.622
            & 0.376
            \\
            B2
            &
            &
            &
            & \checkmark
            &
            &
            & 24.40
            & \textbf{35.07}
            & \textbf{3.672}
            & \textbf{2.883}
            & 19.66
            & 16.75
            & 42.52
            & 33.08
            & 0.613
            & 0.396
            \\
            B3
            &
            &
            & \checkmark
            & \checkmark
            &
            &
            & 28.20
            & 34.59
            & 3.789
            & 2.991
            & 21.02
            & 16.64
            & 44.43
            & 31.93
            & 0.618
            & 0.393
            \\\midrule
            C1
            & \checkmark
            &
            & \checkmark
            &
            &
            &
            & 27.54
            & 19.36
            & 3.992
            & 4.438
            & 21.15
            & 10.12
            & 44.06
            & 24.13
            & 0.619
            & 0.374
            \\
            C2
            &
            & \checkmark
            &
            & \checkmark
            &
            &
            & 24.50
            & 34.91
            & 3.702
            & 2.998
            & 19.79
            & \textbf{17.26}
            & 42.62
            & \textbf{33.36}
            & 0.612
            & \textbf{0.399}
            \\
            C3
            & \checkmark
            & \checkmark
            & \checkmark
            & \checkmark
            &
            &
            & 28.43
            & 33.72
            & 3.779
            & 3.120
            & 21.41
            & 16.32
            & \textbf{45.06}
            & 32.49
            & 0.617
            & 0.396
            \\\midrule
            D1
            &
            &
            & \checkmark
            & \checkmark
            & \checkmark
            & \checkmark
            & \textbf{28.62}
            & 34.55
            & 3.734
            & 2.928
            & \textbf{22.30}
            & 16.48
            & 45.00
            & 32.20
            & 0.617
            & 0.390
            \\
            D2
            & \checkmark
            & \checkmark
            & \checkmark
            & \checkmark
            & \checkmark
            & \checkmark
            & 28.26
            & 34.94
            & 3.761
            & 3.048
            & 21.47
            & 16.49
            & 44.66
            & 32.32
            & \textbf{0.620}
            & 0.392
            \\\bottomrule
        \end{NiceTabularX}
    }
    \label{tbl:losses}
\end{table*}

Our work introduces three novel loss terms for the task of aligning videos to step-by-step instructions.
We now analyze the effectiveness of each loss by evaluating our model trained using different combinations.
The results are summarized in~\tabref{tbl:losses}.
We can draw several conclusions from these results
First, our video-diagram contrastive loss (A) slightly outperforms the standard infoNCE loss used by CLIP.
This confirms our intuition that infoNCE is adversely affected by the many-to-one matchings between video clips and instructional diagrams albeit only slightly.

Second, the video-manual contrastive loss (B) gives the greatest boost in performance over the baseline approaches and once used gain little benefit from the video-diagram contrastive loss (A).
The intra-manual contrastive loss (C) combined with the other losses slightly improves results.

Last, including losses on page diagrams even when evaluating on step diagrams improves results (but not vice versa).
We hypothesize that this is because page diagrams provide a regularizing effect on learning since it is easier to match against pages than individual steps.

\section{Conclusion}
\label{sec:conclusion}

In this paper, we investigated the problem of aligning instructional videos with a high-level schematic representation of the task, depicted by abstract instructional diagrams showing the steps in the process.
We proposed a method based on contrastive learning to align video and diagram features using three novel losses designed specifically for this task.
Our focus is on Ikea furniture assembly where alignment is done between in-the-wild videos and the corresponding official assembly manuals.
To this end, we also collected a dataset of 183 hours of in-the-wild assembly videos and nearly 8,300 diagrams.
Two tasks are designed on this dataset to evaluate the performance of our method: (i) a nearest neighbor retrieval task between video clips and instructional diagrams, (ii) alignment of the instruction diagrams to their corresponding assembly video clips.
On both tasks, experimental results show that our proposed sinusoidal progress rate feature and optimal transport modules lead to better temporal alignment and each one of the proposed losses enables the model to learn better representations, compared with compelling alternatives that do not take into account the unique nature of the problem.

Our work suggests several directions for future work.
First, it would be interesting to consider including additional modalities such as video narrations into our framework.
Second, extending the task to unsupervised or weakly supervised settings would overcome our current limitation of requiring ground truth alignments for learning.
Last, an ambitious long-term goal is to develop applications, built on our alignment model, that automatically monitor and guide a user through an assembly process or facilitate robot-human collaboration on instructional tasks.

\section{Acknowledgement}

This project has received funding from the European Union’s Horizon 2020 research and innovation programme under the Marie Sklodowska-Curie grant agreement No.~893465. This project is also supported by an ARC Future Fellowship No.~FT200100421 and an ANU-MERL PhD scholarship agreement.

{\small
    \bibliographystyle{ieee_fullname}
    \bibliography{paper}
}

\end{document}


\title{Aligning Step-by-Step Instructional Diagrams to Video Demonstrations Supplementary Material}

\author{
Jiahao Zhang$^{1,}$\thanks{Supported by an ANU-MERL PhD scholarship agreement.}\quad
Anoop Cherian$^{2}$\quad
Yanbin Liu$^{1}$\\
Yizhak Ben-Shabat$^{1,3,}$\thanks{Supported by Marie Sklodowska-Curie grant agreement No.~893465.}\quad
Cristian Rodriguez$^{4}$\quad
Stephen Gould$^{1,}$\thanks{Supported by an ARC Future Fellowship No.~FT200100421.}\\
$^1$The Australian National University,
$^2$Mitsubishi Electric Research Labs\\
$^3$Technion Israel Institute of Technology,
$^4$The Australian Institute for Machine Learning\\
{\tt\small $^1$\{first.last\}@anu.edu.au}
{\tt\small $^2$cherian@merl.com}
{\tt\small $^3$sitzikbs@gmail.com}
{\tt\small $^4$crodriguezop@gmail.com}\\
{\tt\small \href{https://davidzhang73.github.io/en/publication/zhang-cvpr-2023/}{https://davidzhang73.github.io/en/publication/zhang-cvpr-2023/}}
}

\maketitle

\appendix

\section{IAW Dataset Statistics}

\begin{figure}[ht]
    \centering
    \includegraphics[width=0.85\linewidth]{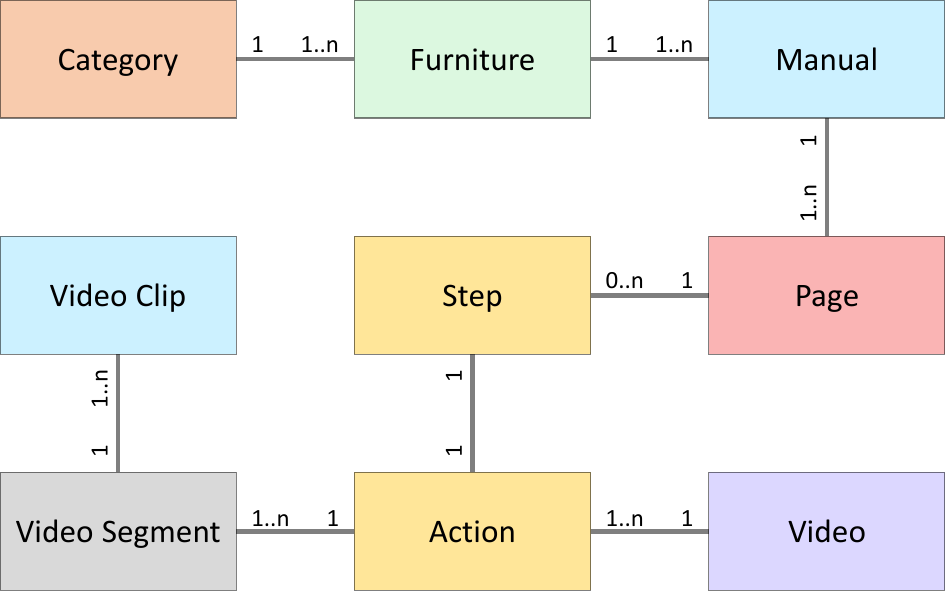}
    \caption{Entity Relationship (ER) Diagram of entities defined for the IAW dataset.}
    \label{fig:er}
\end{figure}

We defined several entities to describe the hierarchical structure of Ikea Assembly in the Wild (IAW) dataset as demonstrated in~\cref{fig:er}.
In terms of the instructional assembly manuals, we have categories, furniture, manuals, pages and steps.
As for videos, there are videos, actions, video segments (10s) and clips (64 frames 2.133s).
A matching is made between a step in the manual and an action from the video, which is a one-to-one relation.
The relation between step and page is many (0..n)-to-one, because a page may contain other information instead of an instructional diagram.
Besides the above two, the rest are all many (1..n)-to-one relation.
It is worth noting that one piece of furniture may correspond to multiple manuals, and we concatenate the manuals according to the assembly order.

In terms of the data collection, we first crawled all manuals under the category \textit{Furniture} from Ikea official website.
We manually found all related assembly videos from YouTube, split the PDF manual into pages and cropped out every individual step diagram. 
With the above information, we out-sourced video to diagram alignment tasks to Amazon Mechanical Turker platform.
The alignments were then audited and refined carefully.
Final statistics are shown in \cref{tbl:manual} and \cref{tbl:video}.

\begin{table}[ht]
    \setlength\extrarowheight{0.5pt}
    \caption{Statistics of assembly manuals categorized by each furniture category.}
    \resizebox{\linewidth}{!}{
        \begin{tabular}{ l c c c c c }
            \toprule
            Category
            & \#furniture
            & \#manuals
            & \#pages
            & \#steps
            \\\midrule
            Beds
            & 33
            & 37
            & 769
            & 823
            \\
            Bookcases \& shelving units
            & 55
            & 61
            & 961
            & 1152
            \\
            Cabinets \& cupboards
            & 28
            & 36
            & 851
            & 920
            \\
            Chairs
            & 74
            & 77
            & 632
            & 884
            \\
            Chests of drawers \& drawer units
            & 35
            & 50
            & 1288
            & 1194
            \\
            Children's furniture
            & 5
            & 5
            & 80
            & 84
            \\
            Furniture sets
            & 2
            & 2
            & 11
            & 20
            \\
            Gaming furniture
            & 1
            & 1
            & 32
            & 24
            \\
            Outdoor furniture
            & 11
            & 11
            & 79
            & 88
            \\
            Sofas
            & 28
            & 31
            & 500
            & 540
            \\
            TV \& media furniture
            & 11
            & 11
            & 321
            & 300
            \\
            Tables \& desks
            & 117
            & 119
            & 2129
            & 1947
            \\
            Trolleys
            & 3
            & 3
            & 48
            & 40
            \\
            Wardrobes
            & 17
            & 17
            & 562
            & 552
            \\\midrule
            Total
            & 420
            & 461
            & 8263
            & 8568
            \\
            Per Category Median
            & 22.5
            & 24.0
            & 531.0
            & 546.0
            \\
            Per Category Average
            & 30.0
            & 32.9
            & 590.2
            & 612.0
            \\\bottomrule
        \end{tabular}
    }
    \label{tbl:manual}
\end{table}

\begin{table}[ht]
    \setlength\extrarowheight{2pt}
    \caption{Statistics of assembly videos categorized by each furniture category. Annotated duration denotes the total duration of video segments that have labels.}
    \resizebox{\linewidth}{!}{
        \begin{tabular}{ m{20ex} c c c c c }
            \toprule
            Category
            & \#videos
            & \#actions
            & Duration
            & Annotated Duration
            \\\midrule
            Beds
            & 86
            & 1554
            & 16h:43m:00s
            & 10h:50m:59s
            \\
            Bookcases \& shelving units
            & 128
            & 1636
            & 21h:23m:55s
            & 12h:42m:08s
            \\
            Cabinets \& cupboards
            & 49
            & 1082
            & 11h:29m:12s
            & 07h:12m:38s
            \\
            Chairs
            & 185
            & 1478
            & 21h:22m:27s
            & 11h:44m:13s
            \\
            Chests of drawers \& drawer units
            & 113
            & 2891
            & 30h:51m:36s
            & 20h:40m:37s
            \\
            Children's furniture
            & 7
            & 52
            & 01h:14m:37s
            & 00h:32m:27s
            \\
            Furniture sets
            & 3
            & 15
            & 00h:17m:19s
            & 00h:02m:36s
            \\
            Gaming furniture
            & 1
            & 31
            & 00h:00m:56s
            & 00h:00m:42s
            \\
            Outdoor furniture
            & 11
            & 72
            & 01h:25m:08s
            & 00h:37m:57s
            \\
            Sofas
            & 83
            & 1285
            & 15h:08m:20s
            & 08h:44m:44s
            \\
            TV \& media furniture
            & 36
            & 749
            & 07h:41m:47s
            & 05h:31m:39s
            \\
            Tables \& desks
            & 266
            & 3947
            & 47h:07m:37s
            & 29h:26m:20s
            \\
            Trolleys
            & 8
            & 79
            & 00h:51m:59s
            & 00h:34m:14s
            \\
            Wardrobes
            & 31
            & 812
            & 07h:29m:12s
            & 05h:02m:30s
            \\\midrule
            Total
            & 1007
            & 15683
            & 183h:07m:05s
            & 113h:43m:48s
            \\
            Per Category Median
            & 42.5
            & 947.0
            & 09h:35m:29s
            & 06h:22m:08s
            \\
            Per Category Average
            & 71.9
            & 1120.2
            & 13h:04m:47s
            & 08h:07m:24s
            \\\bottomrule
        \end{tabular}
    }
    \label{tbl:video}
\end{table}

\begin{figure}[ht]
    \centering
    \includegraphics[width=0.9\linewidth]{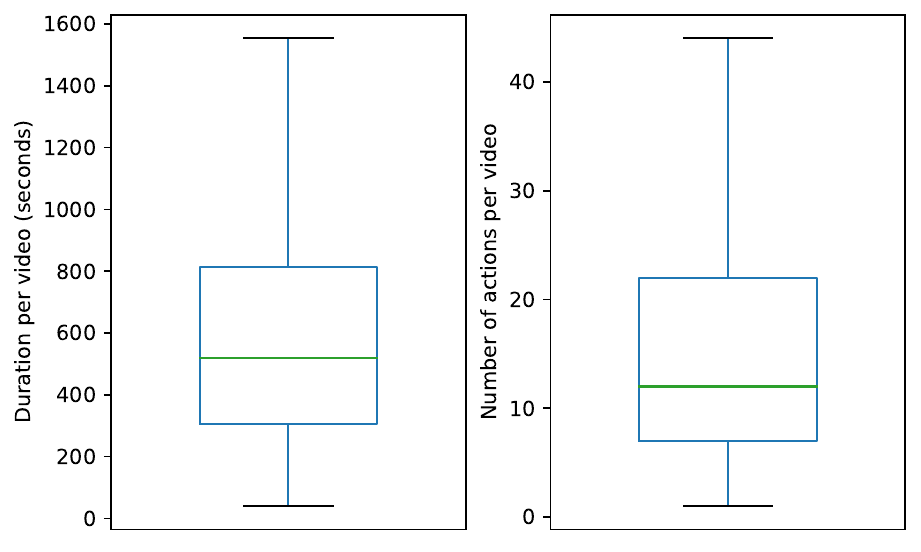}
    \caption{Distribution of video duration and the number of actions per video. Video duration: (max: 4763, min: 40, mean: 655, median: 520); Number of actions: (max: 63, min: 1, mean: 16, median: 12)}
    \label{fig:boxplot}
\end{figure}

The distributions of video duration and the number of actions per video are shown in the boxplot below~(\cref{fig:boxplot}).

\begin{figure}[ht]
    \centering
    \begin{subfigure}{0.49\linewidth}
        \includegraphics[width=\linewidth]{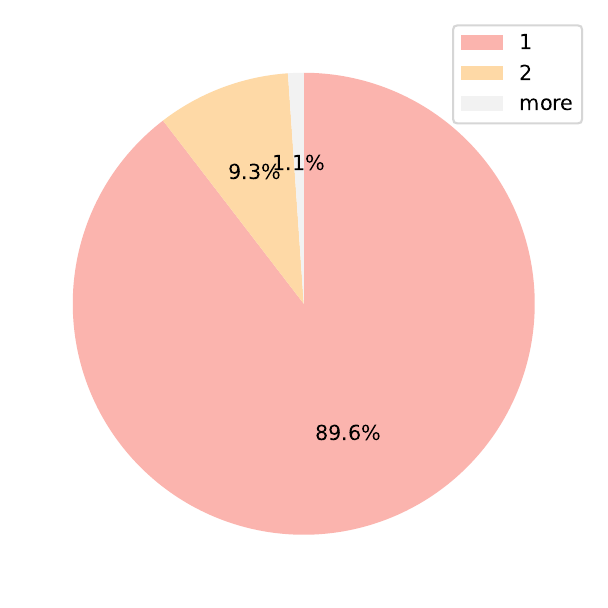}
        \caption{How many people are involved during assembling.}
        \label{fig:video_people_count}
    \end{subfigure}
    \begin{subfigure}{0.49\linewidth}
        \includegraphics[width=\linewidth]{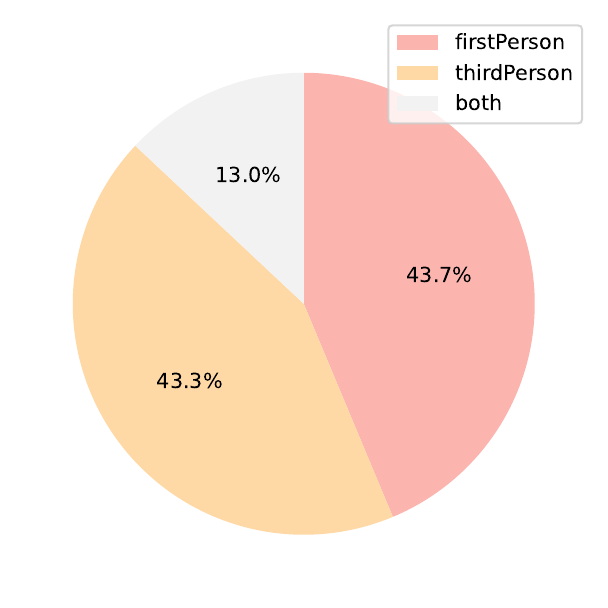}
        \caption{Is it first-person view, third-person view or both occurred.}
        \label{fig:video_person_view}
    \end{subfigure}
    \begin{subfigure}{0.49\linewidth}
        \includegraphics[width=\linewidth]{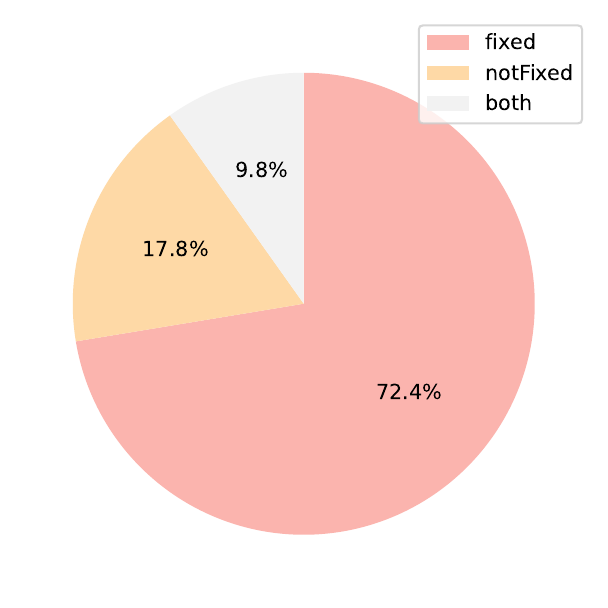}
        \caption{Is the camera fixed, not-fixed or both occurred.}
        \label{fig:video_is_fixed}
    \end{subfigure}
    \begin{subfigure}{0.49\linewidth}
        \includegraphics[width=\linewidth]{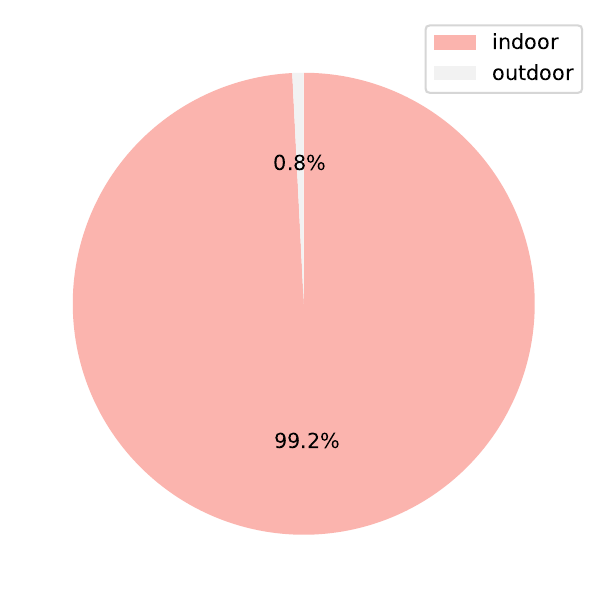}
        \caption{Is the assembly happened indoor or outdoor.}
        \label{fig:video_is_indoor}
    \end{subfigure}
    \caption{Proportion of the four attributes.}
    \label{fig:video_collect}
\end{figure}

\begin{figure}[t]
    \centering
    \includegraphics[width=0.95\linewidth]{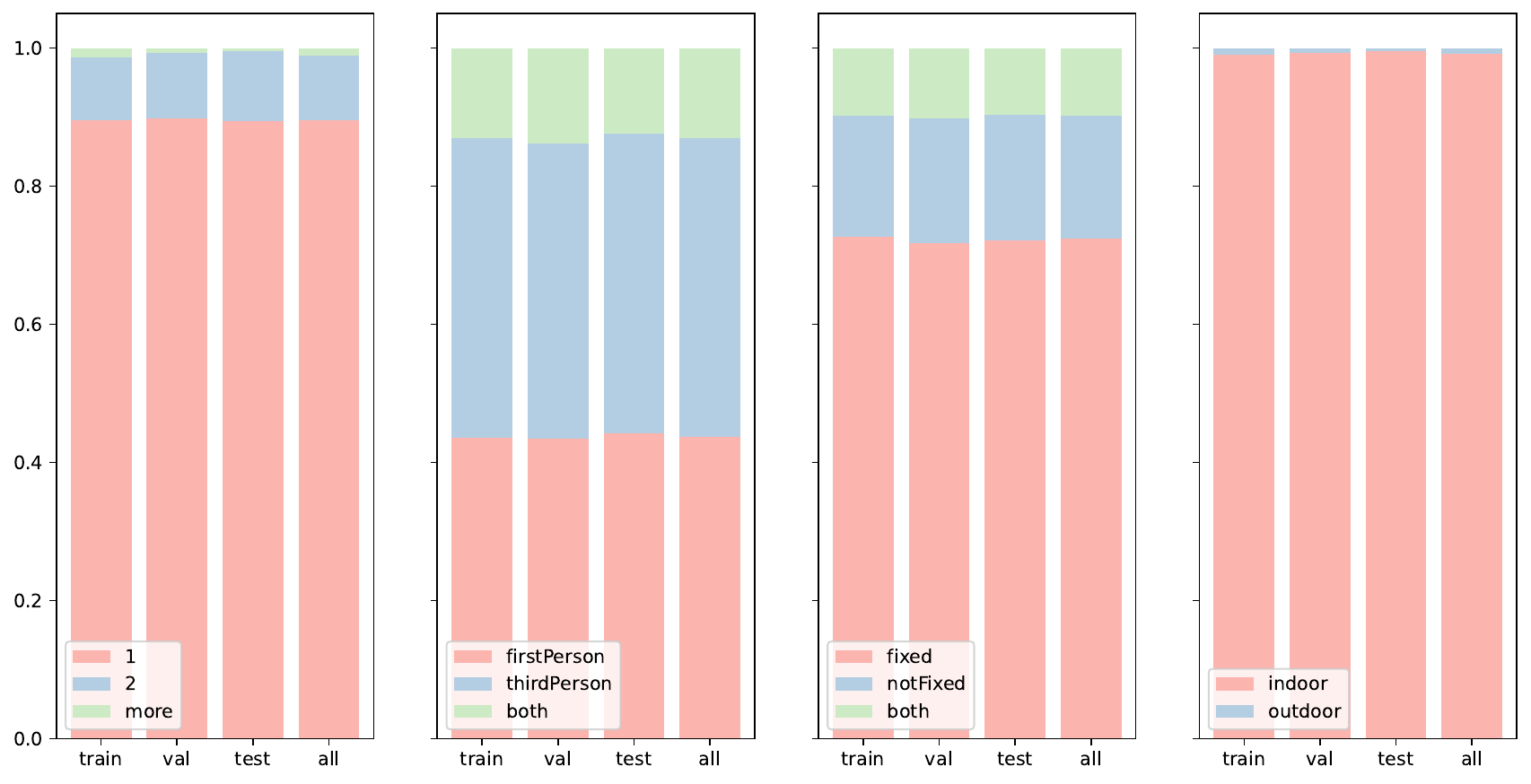}
    \caption{Balanced distribution of four attributes in each split comparing with the entire dataset (all).}
    \label{fig:split}
\end{figure}

As shown in~\cref{fig:video_collect}, we manually attached four attributes for each video.
When splitting the IAW dataset into train, validation and test splits, a greedy algorithm is used to balance the distribution in each split w.r.t. four attributes as shown in~\cref{fig:split}.
Concretely, the greedy algorithm traverses each furniture. If a furniture contains only one video, then the video is added to train split; if two videos, then one for train and one for test; if more than two videos, one for test, one for validation and put all the rest into the train split. 
We try every possible assignments of videos in this situation, and select the one that minimises the distribution difference between the split and the entire dataset. 
In this way, we can ensure that there is no shared video between train and validation or test split and all manuals are fed to the model during training. 
The final split statistics are shown in the~\cref{tbl:split}.

\begin{table}[t]
    \centering
    \small
    \begin{tabular}{ l c c c c }
        \toprule
        & Train
        & Validation
        & Test
        \\\midrule
        \#furniture
        & 420
        & 138
        & 226
        \\
        \#videos
        & 643
        & 138
        & 226
        \\
        \#actions
        & 9925
        & 2221
        & 3537
        \\
        \#video segments
        & 30876
        & 6871
        & 11103
        \\
        \bottomrule
    \end{tabular}
    \caption{Split Statistics.}
    \label{tbl:split}
\end{table}

\textbf{We will release the dataset including video URLs and full annotations on publication of this paper.} 
In addition, we will also make public the tool we used for annotating and the supporting infrastructures we developed for Amazon Mechanical Turker.

\section{Experimental Results}

In this section, we provide more experimental results both quantitatively and qualitatively to demonstrate the effectiveness of our method. 

\subsection{Quantitative Results}

\begin{table*}[ht]
    \centering
    \small
    \caption{Ablation study on SPRF. These experiments are conducted based on loss configuration B3.}
    \newcolumntype{Z}{>{\centering\arraybackslash}X}
    \resizebox{\linewidth}{!}{
        \begin{NiceTabular}{
                @{} 
                w{l}{5em}
                *{4}{Z} 
                | 
                *{6}{Z} 
            }
            \toprule
            \Block{3-1}{Method}
            & \Block{1-4}{Video to diagram retrieval}
            &
            &
            &
            & \Block{1-6}{Diagram to video retrieval}
            \\
            & \Block{1-2}{Top1 Acc.\%$\uparrow$}
            &
            & \Block{1-2}{AIE$\downarrow$}
            &
            & \Block{1-2}{R@1$\uparrow$}
            &
            & \Block{1-2}{R@3$\uparrow$}
            &
            & \Block{1-2}{AUROC$\uparrow$}
            \\
            \cmidrule(lr){2-3}\cmidrule(lr){4-5}\cmidrule(lr){6-7}\cmidrule(lr){8-9}\cmidrule(lr){10-11}
            & S
            & P
            & S
            & P
            & S
            & P
            & S
            & P
            & S
            & P
            \\\midrule
            w/o
            & 22.28
            & 27.70
            & 5.983
            & 4.639
            & 16.97
            & 12.95
            & 36.36
            & 27.25
            & 0.548
            & 0.357
            \\\midrule
            PE\cite{vaswani2017attention} Add
            & 19.10
            & 25.72
            & 4.317
            & 3.248
            & 14.49
            & 12.71
            & 35.04
            & 26.93
            & 0.544
            & 0.356
            \\
            PE\cite{vaswani2017attention} Concat
            & 18.85
            & 24.93
            & 4.384
            & 3.265
            & 15.28
            & 12.39
            & 34.25
            & 27.04
            & 0.541
            & 0.353
            \\\midrule
            PRF
            & 27.29
            & 32.60
            & 3.830
            & 3.128
            & \textbf{21.08}
            & 16.09
            & 43.89
            & 31.03
            & 0.615
            & 0.393
            \\
            SPRF After
            & 25.75
            & 34.17
            & \textbf{3.594}
            & 3.144
            & 20.08
            & 16.50
            & 43.09
            & 31.78
            & 0.617
            & \textbf{0.394}
            \\
            SPRF
            & \textbf{28.20}
            & \textbf{34.59}
            & 3.789
            & \textbf{2.991}
            & 21.02
            & \textbf{16.64}
            & \textbf{44.43}
            & \textbf{31.93}
            & \textbf{0.618}
            & 0.393
            \\\bottomrule
        \end{NiceTabular}
    }
    \label{tbl:sprf}
\end{table*}

\begin{table*}[ht]
    \centering
    \small
    \caption{Ablation study for different batch sizes based on CLIP and A3 without OT.}
    \newcolumntype{Z}{>{\centering\arraybackslash}X}
    \resizebox{\linewidth}{!}{
        \begin{NiceTabular}{
                @{}
                X
                *{5}{Z} 
                |
                *{6}{Z}
            }
            \toprule
            \Block{3-1}{Method}
            & \Block{3-1}{Batch Size}
            & \Block{1-4}{Video to diagram retrieval}
            &
            &
            &
            & \Block{1-6}{Diagram to video retrieval}
            \\
            &
            & \Block{1-2}{Top1 Acc.\%$\uparrow$}
            &
            & \Block{1-2}{AIE$\downarrow$}
            &
            & \Block{1-2}{R@1$\uparrow$}
            &
            & \Block{1-2}{R@3$\uparrow$}
            &
            & \Block{1-2}{AUROC$\uparrow$}
            \\
            \cmidrule(lr){3-4}\cmidrule(lr){5-6}\cmidrule(lr){7-8}\cmidrule(lr){9-10}\cmidrule(lr){11-12}
            & 
            & S
            & P
            & S
            & P
            & S
            & P
            & S
            & P
            & S
            & P
            \\\midrule
            \multirow{3}*{CLIP}
            & 64
            & 22.59
            & 23.10
            & 4.011
            & 3.976
            & 19.74
            & 12.07
            & 42.43
            & 27.43
            & 0.611
            & 0.386
            \\
            & 128
            & 22.08
            & 23.67
            & 4.186
            & 3.870
            & 18.71
            & 12.17
            & 41.64
            & 27.12
            & 0.606
            & 0.387
            \\
            & 256
            & 19.61
            & 19.05
            & 4.274
            & 4.180
            & 16.94
            & 10.25
            & 38.67
            & 23.45
            & 0.590
            & 0.373
            \\\midrule
            \multirow{3}*{A3}
            & 64
            & 22.18
            & 23.28
            & 4.097
            & 3.972
            & 19.48
            & 12.63
            & 42.58
            & 27.13
            & 0.610
            & 0.387
            \\
            & 128
            & 21.71
            & 22.84
            & 3.999
            & 3.956
            & 19.73
            & 12.74
            & 40.97
            & 27.42
            & 0.601
            & 0.383
            \\
            & 256
            & 20.58
            & 19.34
            & 4.036
            & 4.090
            & 17.08
            & 10.13
            & 39.89
            & 24.64
            & 0.583
            & 0.371
            \\\bottomrule
        \end{NiceTabular}
    }
    \label{tbl:bs}
\end{table*}

\begin{table*}[ht]
    \centering
    \small
    \caption{Ablation study results on different loss combinations. All the results in this table are obtained \textbf{after} applying the optimal transport post-processing. Optimal transport  is generally effective for most model variants.} 
    \newcolumntype{Z}{>{\centering\arraybackslash}X}
    \resizebox{\linewidth}{!}{
        \begin{NiceTabular}{
                @{}
                w{l}{2.5em}
                *{6}{w{l}{0.4em}}
                *{4}{Z}
                |
                *{6}{Z}
            }
            \toprule
            \Block{3-1}{Exp.}
            &
            &
            &
            &
            &
            &
            & \Block{1-4}{Video to diagram retrieval}
            &
            &
            &
            & \Block{1-6}{Diagram to video retrieval}
            \\
            & \Block{1-2}{Loss A}
            &
            & \Block{1-2}{Loss B}
            &
            & \Block{1-2}{Loss C}
            &
            & \Block{1-2}{Top1 Acc.\%$\uparrow$}
            &
            & \Block{1-2}{AIE$\downarrow$}
            &
            & \Block{1-2}{R@1$\uparrow$}
            &
            & \Block{1-2}{R@3$\uparrow$}
            &
            & \Block{1-2}{AUROC$\uparrow$}
            \\
            \cmidrule(lr){2-3}\cmidrule(lr){4-5}\cmidrule(lr){6-7}\cmidrule(lr){8-9}\cmidrule(lr){10-11}\cmidrule(lr){12-13}\cmidrule(lr){14-15}\cmidrule(lr){16-17}
            & S
            & P
            & S
            & P
            & S
            & P
            & S
            & P
            & S
            & P
            & S
            & P
            & S
            & P
            & S
            & P
            \\\midrule
            CosSim\textsuperscript{\dag}
            &
            &
            &
            &
            &
            &
            & 17.47
            & 9.97
            & 3.837
            & 4.802
            & 15.70
            & 6.50
            & 39.78
            & 18.91
            & 0.574
            & 0.356
            \\
            CLIP\textsuperscript{\dag}
            &
            &
            &
            &
            &
            &
            & 20.52
            & 19.36
            & 4.175
            & 4.104
            & 18.72
            & 10.71
            & 39.35
            & 22.23
            & 0.553
            & 0.352
            \\\midrule
            A1\textsuperscript{\dag}
            & \checkmark
            &
            &
            &
            &
            & 
            & 20.56
            & 14.44
            & 4.323
            & 4.892
            & 18.63
            & 8.42
            & 39.81
            & 18.84
            & 0.547
            & 0.330
            \\
            A2\textsuperscript{\dag}
            &
            & \checkmark
            &
            &
            &
            &
            & 19.17
            & 18.29
            & 4.203
            & 4.420
            & 17.35
            & 10.13
            & 37.30
            & 20.71
            & 0.534
            & 0.342
            \\
            A3\textsuperscript{\dag}
            &
            \checkmark
            & \checkmark
            &
            & 
            & 
            &
            & 20.18
            & 17.90
            & 4.236
            & 4.574
            & 17.48
            & 9.51
            & 38.67
            & 20.62
            & 0.538
            & 0.341
            \\\midrule
            B1
            &
            & 
            & \checkmark
            &
            &
            &
            & 29.54
            & 21.08
            & 3.563
            & 4.134
            & 24.25
            & 11.48
            & 46.64
            & 24.81
            & 0.607
            & 0.369
            \\
            B2
            &
            & 
            &
            & \checkmark
            &
            &
            & 25.99
            & \textbf{36.74}
            & 3.528
            & \textbf{2.791}
            & 22.79
            & 18.33
            & 45.01
            & 31.18
            & 0.596
            & 0.393
            \\
            B3
            &
            & 
            & \checkmark
            & \checkmark
            &
            &
            & 29.74
            & 36.40
            & 3.605
            & 2.880
            & 24.22
            & 17.89
            & 46.71
            & 29.96
            & 0.598
            & 0.389
            \\\midrule
            C1
            & \checkmark
            &
            & \checkmark
            &
            &
            &
            & 29.29
            & 19.61
            & 3.754
            & 4.402
            & 23.59
            & 10.82
            & 45.77
            & 23.43
            & 0.590
            & 0.355
            \\
            C2
            &
            & \checkmark
            &
            & \checkmark
            &
            &
            & 25.67
            & 36.22
            & 3.588
            & 2.890
            & 22.11
            & 18.12
            & 44.67
            & 30.30
            & 0.589
            & 0.393
            \\
            C3
            & \checkmark
            & \checkmark
            & \checkmark
            & \checkmark
            &
            &
            & 30.37
            & 35.49
            & 3.606
            & 3.022
            & 24.04
            & 18.02
            & 46.31
            & 29.44
            & 0.593
            & 0.389
            \\\midrule
            D1
            &
            &
            & \checkmark
            & \checkmark
            & \checkmark
            & \checkmark
            & \textbf{31.61}
            & 36.71
            & \textbf{3.458}
            & 2.816
            & \textbf{26.62}
            & 18.28
            & \textbf{49.11}
            & \textbf{32.28}
            & \textbf{0.626}
            & \textbf{0.401}
            \\
            D2
            &
            \checkmark
            & \checkmark
            & \checkmark
            & \checkmark
            & \checkmark
            & \checkmark
            & 30.66
            & 36.12
            & 3.539
            & 2.939
            & 25.31
            & \textbf{18.44}
            & 48.86
            & 31.32
            & 0.620
            & 0.396
            \\\bottomrule
        \end{NiceTabular}
    }
    \label{tbl:losses_ot}
\end{table*}

\textbf{Sinusoidal Progress Rate Feature (SPRF).} 
Firstly, we conducted ablation experiments on Positional Embedding (PE) proposed by~\cite{radford2021learning}. 
The original PE is used to represent the order information for words in a sentence of arbitrary length.
In our task, we need to align two different progress rates with different scales, so we manually set length for PE to be 100, and sample positional embeddings from it.
We tried to replace the proposed SPRF with either adding (denoted as ``PE Add'') the positional embedding or concatenating (denoted as ``PE Concat'') to the feature. 
As shown in the~\cref{tbl:sprf}, both variants of position embedding are inferior to our baseline. 
Secondly, we tried to modify the overall architecture so that the SPRF locates after the final linear layer and before the loss (denoted as ``SPRF After''), to make the progress rate information applied directly on the contrastive loss.
This modification, however, failed to outperform the proposed architecture. 
We conjecture that it is because the vision and progress rate features can be better fused through a fully connected layer. 
Besides that, we removed sinusoidal transform (denoted as ``PRF'') and also found the performance dropped. 
Converting progress rate feature into sinusoidal space is intuitive because we are using cosine similarity for distance calculation. 

\begin{figure}[t]
    \centering
    \includegraphics[width=\linewidth]{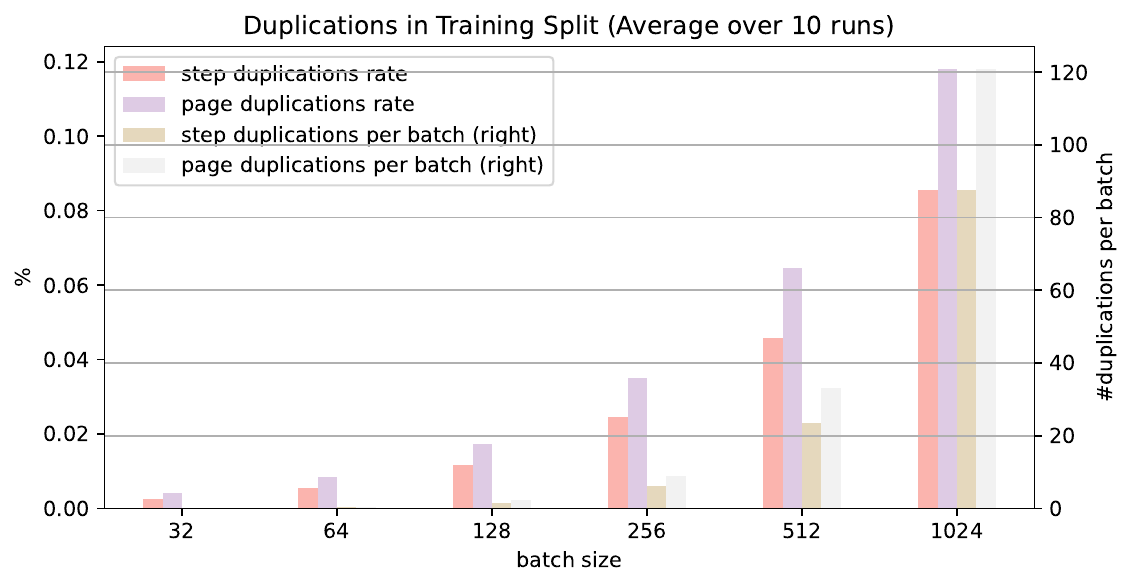}
    \caption{Duplication statistics in a batch w.r.t different batch sizes.}
    \label{fig:duplications}
\end{figure}

\textbf{Batch Size.} Batch size is important for contrastive learning, since the larger batch size leads to more negative samples, hence the stronger supervision signal. 
However, in this task, it failed to increase performance when we enlarge the batch size.
We suspect it is due to the fact that there are semantic collisions or duplications (\cref{fig:duplications}) because there are multiple segments for each action. 
As shown in our ablation study~(\cref{tbl:bs}), all the metrics of both CLIP and A3 drops as batch size increases. 
But A3 has a relatively slower rate of decline compared with CLIP.
Due to the GPU memory constraints, we didn't report results for batch size $>$256.

\textbf{Post Process.} 
We use optimal transport for post-process and the results in \cref{tbl:losses_ot} show that OT improves performance for most cases. 
In particular, for the loss combination D1, it outperforms the one without OT (Tab. 2 in the main paper) by almost 3\%. 
And it is worth noting that compared with B3, Loss C plays a positive role in terms of the performance. 

\subsection{Qualitative Results}

In this section, we show sixteen selected examples: eight for successful alignments and another eight for failure cases. 

Moreover, three video examples are attached in the zip file demonstrating the result of our work. We encourage readers to watch the attached videos for a better understanding of our method and dataset. 

\begin{figure*}[ht]
    \centering
    \begin{subfigure}{0.86\linewidth}
        \includegraphics[width=\linewidth]{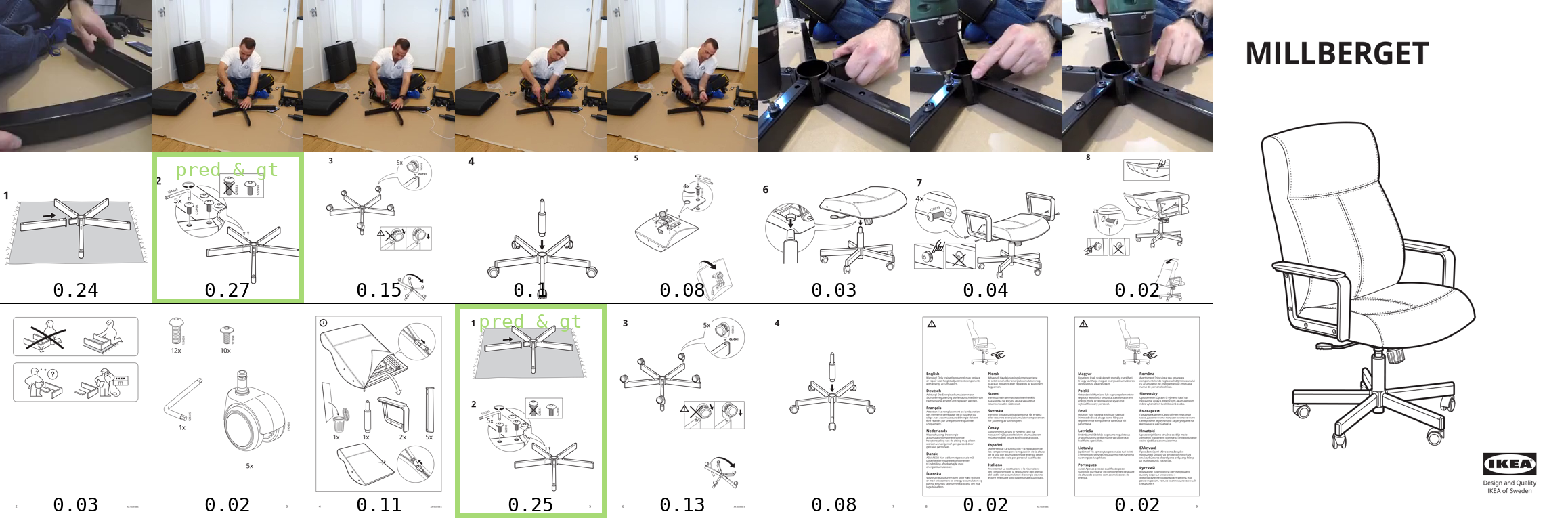}
        \caption{Alignment between a clip from YouTube video \href{https://www.youtube.com/watch?v=36T-ytb8EhM}{36T-ytb8EhM} and an Ikea furniture manual \href{https://www.ikea.com/au/en/search/products/?q=00339416}{00339416}.}
    \end{subfigure}
    \begin{subfigure}{0.86\linewidth}
        \includegraphics[width=\linewidth]{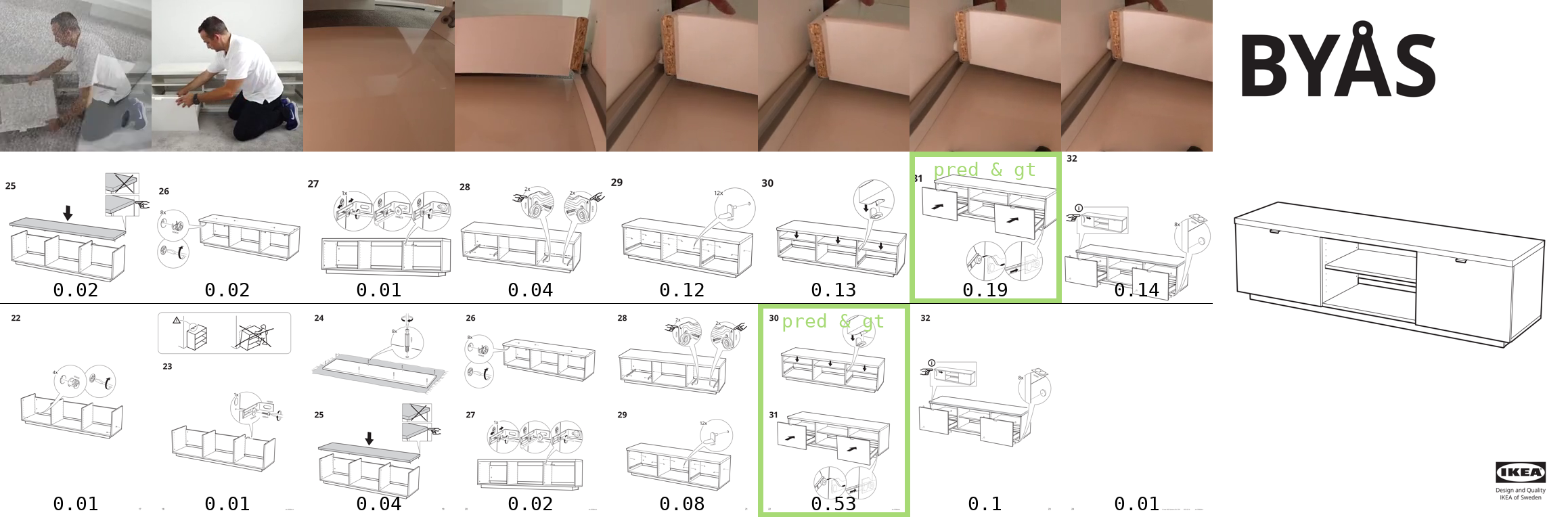}
        \caption{Alignment between a clip from YouTube video \href{https://www.youtube.com/watch?v=aPJF36W93wE}{aPJF36W93wE} and an Ikea furniture manual \href{https://www.ikea.com/au/en/p/byas-tv-bench-high-gloss-white-00352573/}{00352573}.}
    \end{subfigure}
    \begin{subfigure}{0.86\linewidth}
        \includegraphics[width=\linewidth]{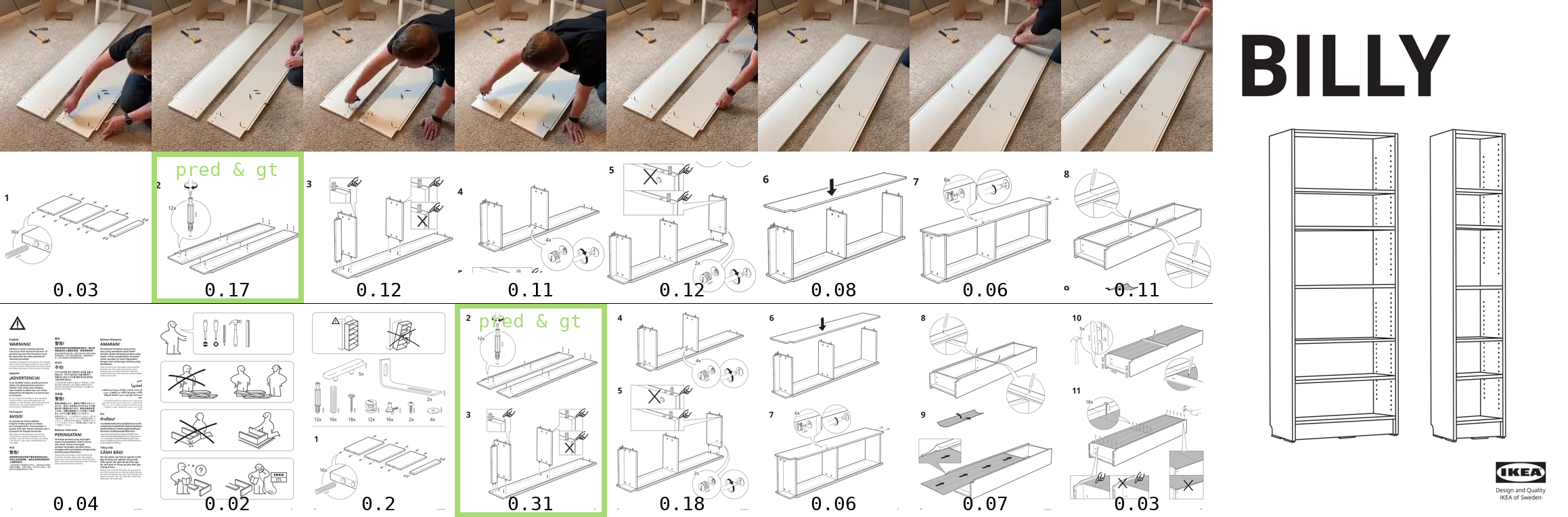}
        \caption{Alignment between a clip from YouTube video \href{https://www.youtube.com/watch?v=Lji-ZnRLPBQ}{Lji-ZnRLPBQ} and an Ikea furniture manual \href{https://www.ikea.com/au/en/p/billy-bookcase-white-10351568/}{10351568}.}
    \end{subfigure}
    \begin{subfigure}{0.86\linewidth}
        \includegraphics[width=\linewidth]{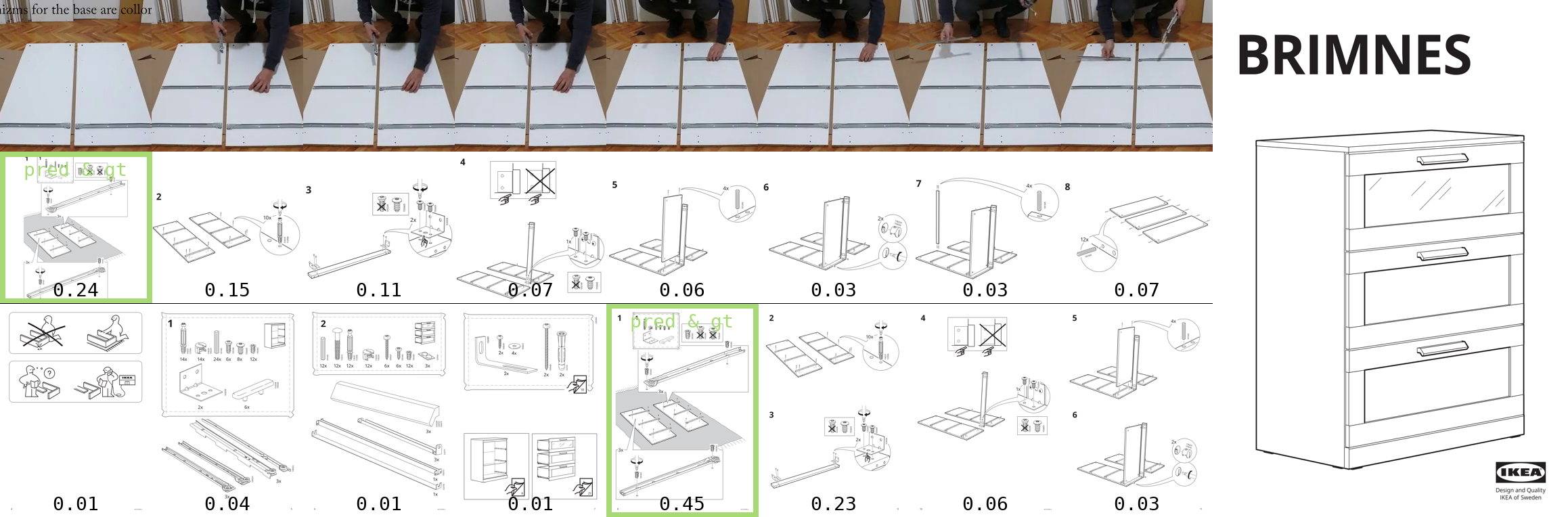}
        \caption{Alignment between a clip from YouTube video \href{https://www.youtube.com/watch?v=n7ZlhsVKnuY}{n7ZlhsVKnuY} and an Ikea furniture manual \href{https://www.ikea.com/au/en/p/brimnes-chest-of-3-drawers-white-frosted-glass-10355415/}{10355415}.}
    \end{subfigure}
    \caption{Eight success examples (4/8).}
    \label{fig:examples1}
\end{figure*}

\setcounter{figure}{5}

\begin{figure*}[ht]
    \centering
    \begin{subfigure}{0.86\linewidth}
        \addtocounter{subfigure}{4}
        \includegraphics[width=\linewidth]{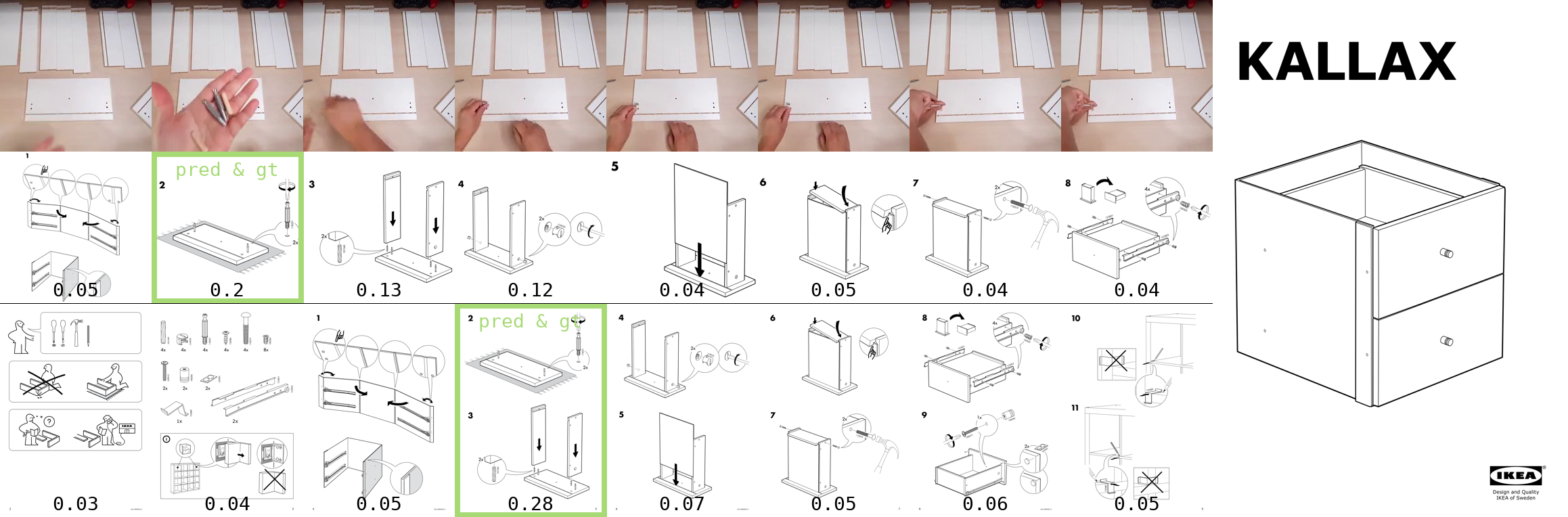}
        \caption{Alignment between a clip from YouTube video \href{https://www.youtube.com/watch?v=G3LS_4PRlEs}{G3LS\_4PRlEs} and an Ikea furniture manual \href{https://www.ikea.com/au/en/p/kallax-insert-with-2-drawers-white-20351879/}{20351879}.}
    \end{subfigure}
    \begin{subfigure}{0.86\linewidth}
        \includegraphics[width=\linewidth]{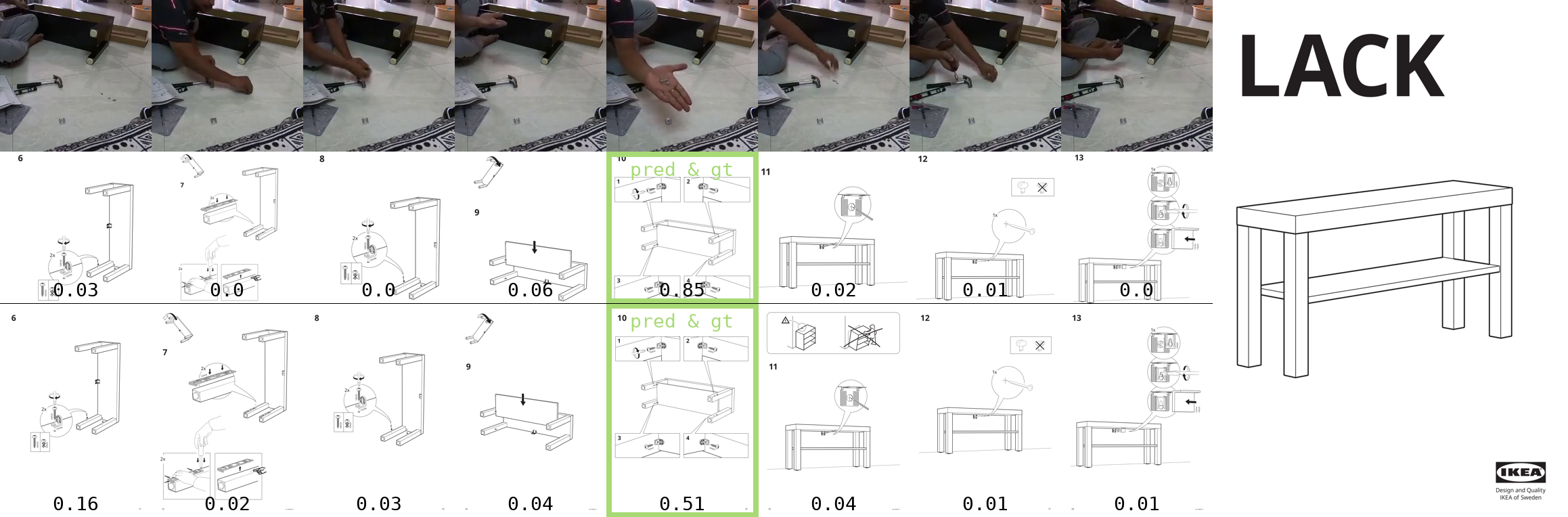}
        \caption{Alignment between a clip from YouTube video \href{https://www.youtube.com/watch?v=mr92aJSM2hM}{mr92aJSM2hM} and an Ikea furniture manual \href{https://www.ikea.com/au/en/p/lack-tv-bench-black-30353566/}{30353566}.}
    \end{subfigure}
    \begin{subfigure}{0.86\linewidth}
        \includegraphics[width=\linewidth]{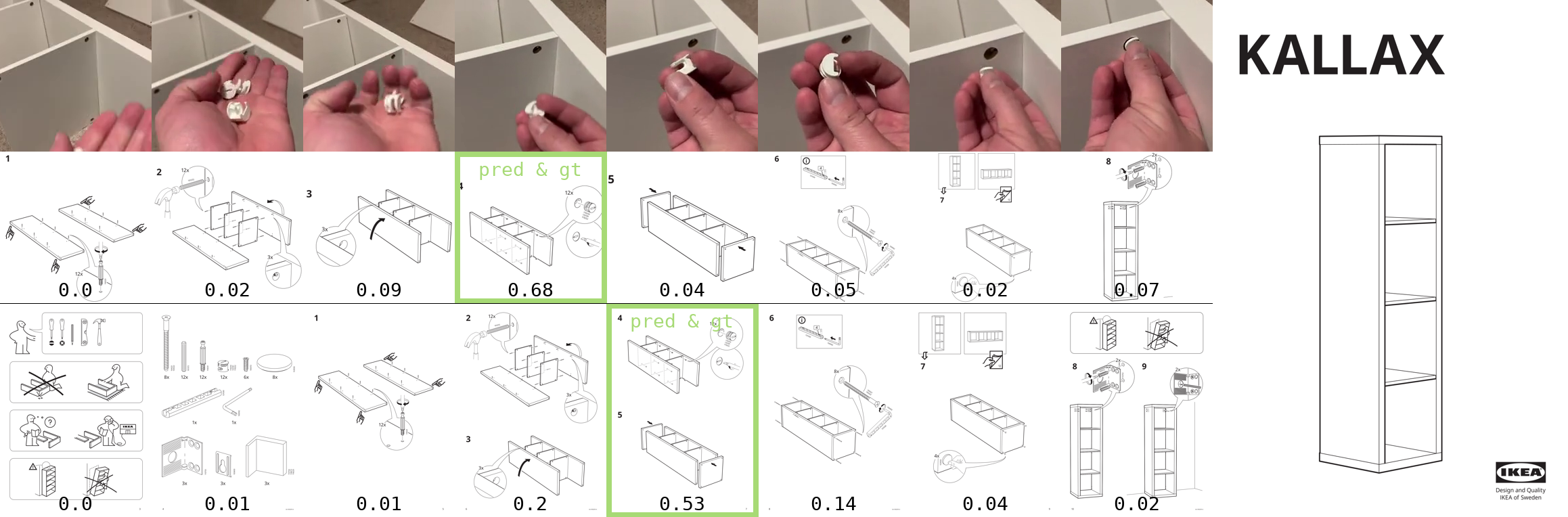}
        \caption{Alignment between a clip from YouTube video \href{https://www.youtube.com/watch?v=jthbbffqPlg}{jthbbffqPlg} and an Ikea furniture manual \href{https://www.ikea.com/au/en/p/kallax-shelving-unit-white-40351883/}{40351883}.}
    \end{subfigure}
    \begin{subfigure}{0.86\linewidth}
        \includegraphics[width=\linewidth]{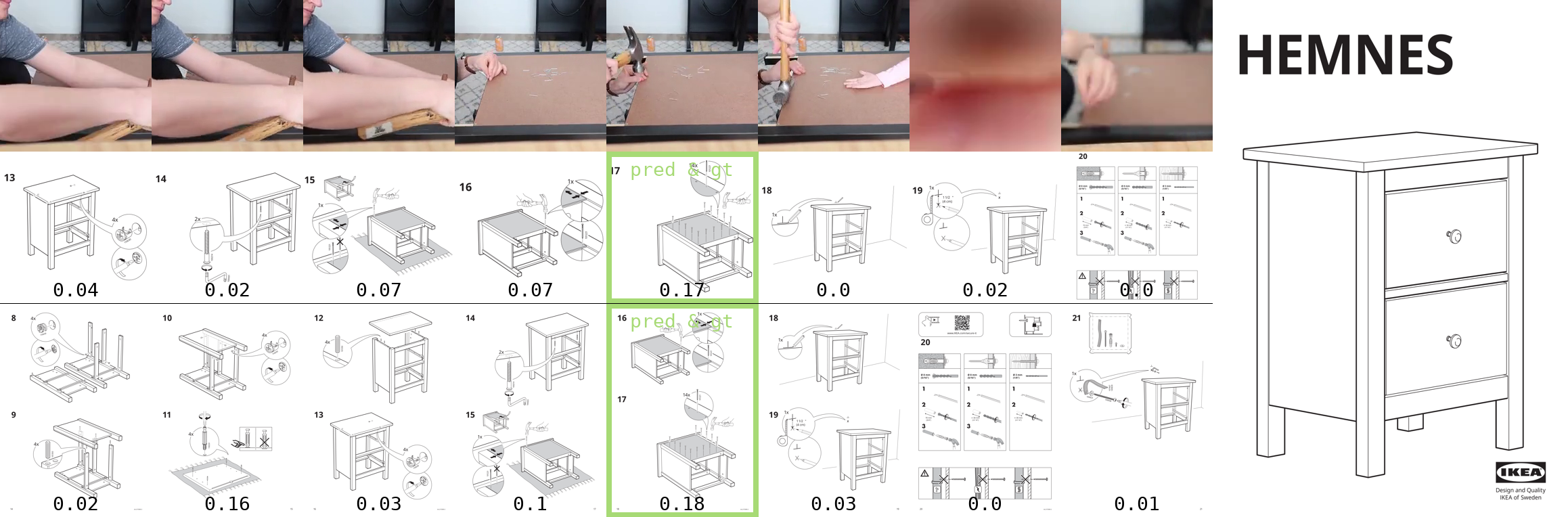}
        \caption{Alignment between a clip from YouTube video \href{https://www.youtube.com/watch?v=n7ZlhsVKnuY}{n7ZlhsVKnuY} and an Ikea furniture manual \href{https://www.ikea.com/au/en/p/hemnes-chest-of-2-drawers-white-stain-50355687/}{50355687}.}
    \end{subfigure}
    \caption{Eight success examples (8/8).}
    \label{fig:examples2}
\end{figure*}

\begin{figure*}[ht]
    \centering
    \begin{subfigure}{0.86\linewidth}
        \includegraphics[width=\linewidth]{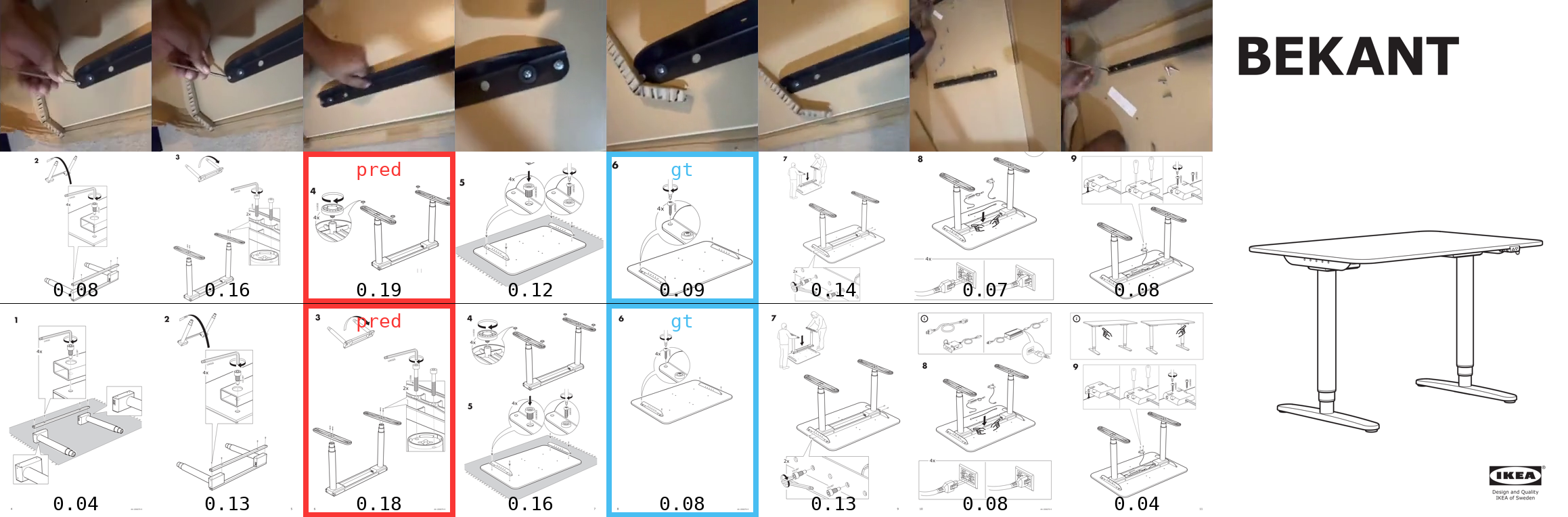}
        \caption{Alignment between a clip from YouTube video \href{https://www.youtube.com/watch?v=TWVmJ3f1U8A}{TWVmJ3f1U8A} and an Ikea furniture manual \href{https://www.ikea.com/au/en/p/bekant-desk-sit-stand-white-stained-oak-veneer-white-s79282252/}{s79282252}.}
    \end{subfigure}
    \begin{subfigure}{0.86\linewidth}
        \includegraphics[width=\linewidth]{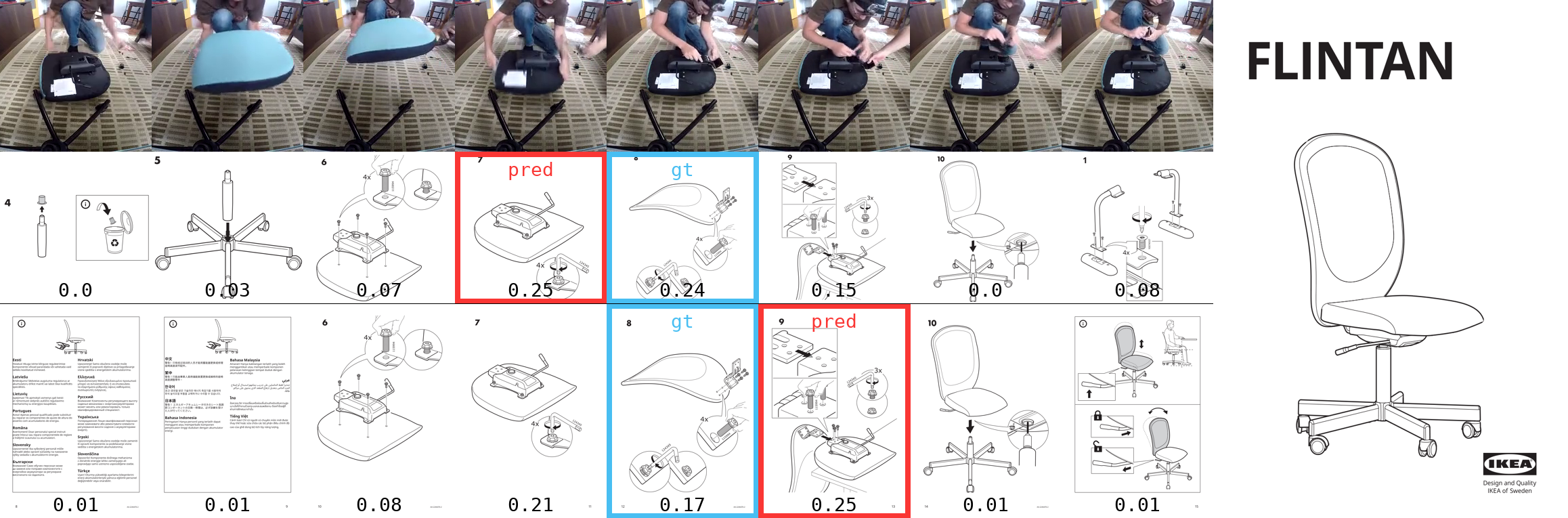}
        \caption{Alignment between a clip from YouTube video \href{https://www.youtube.com/watch?v=CM-n5RaTtOw}{CM-n5RaTtOw} and an Ikea furniture manual \href{https://www.ikea.com/au/en/p/flintan-office-chair-with-armrests-black-s69424469/}{s69424469}.}
    \end{subfigure}
    \begin{subfigure}{0.86\linewidth}
        \includegraphics[width=\linewidth]{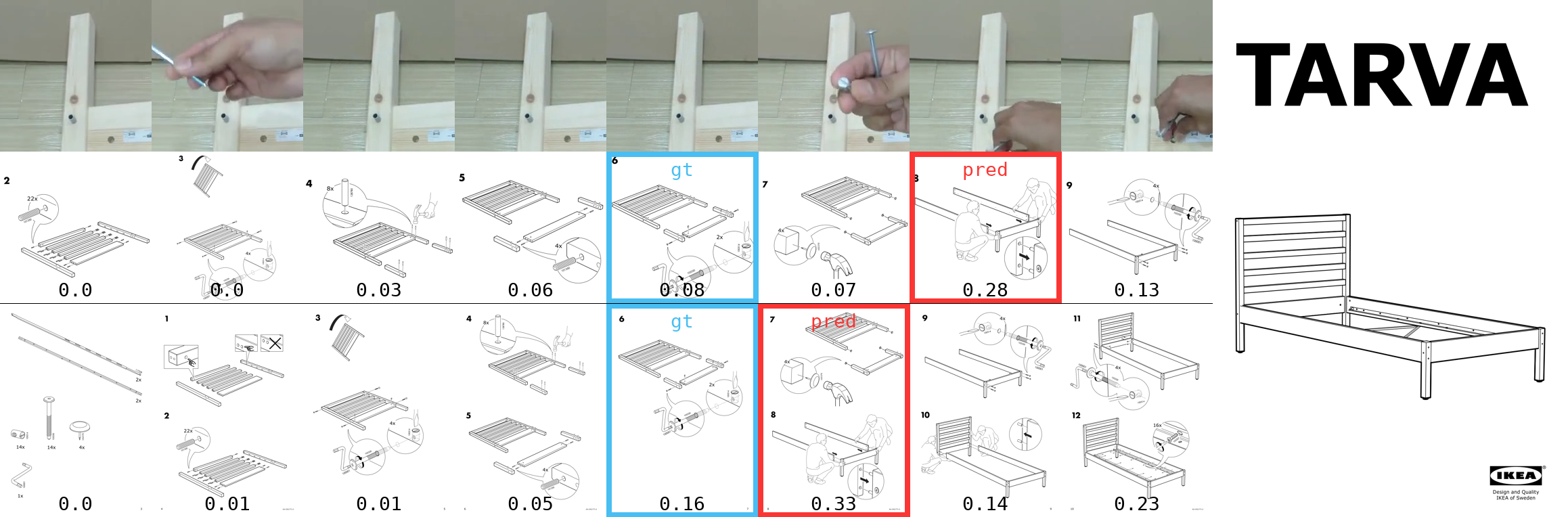}
        \caption{Alignment between a clip from YouTube video \href{https://www.youtube.com/watch?v=u-2oTXnp48c}{u-2oTXnp48c} and an Ikea furniture manual \href{https://www.ikea.com/au/en/p/tarva-bed-frame-pine-luroey-s09009572/}{s09009572}.}
    \end{subfigure}
    \begin{subfigure}{0.86\linewidth}
        \includegraphics[width=\linewidth]{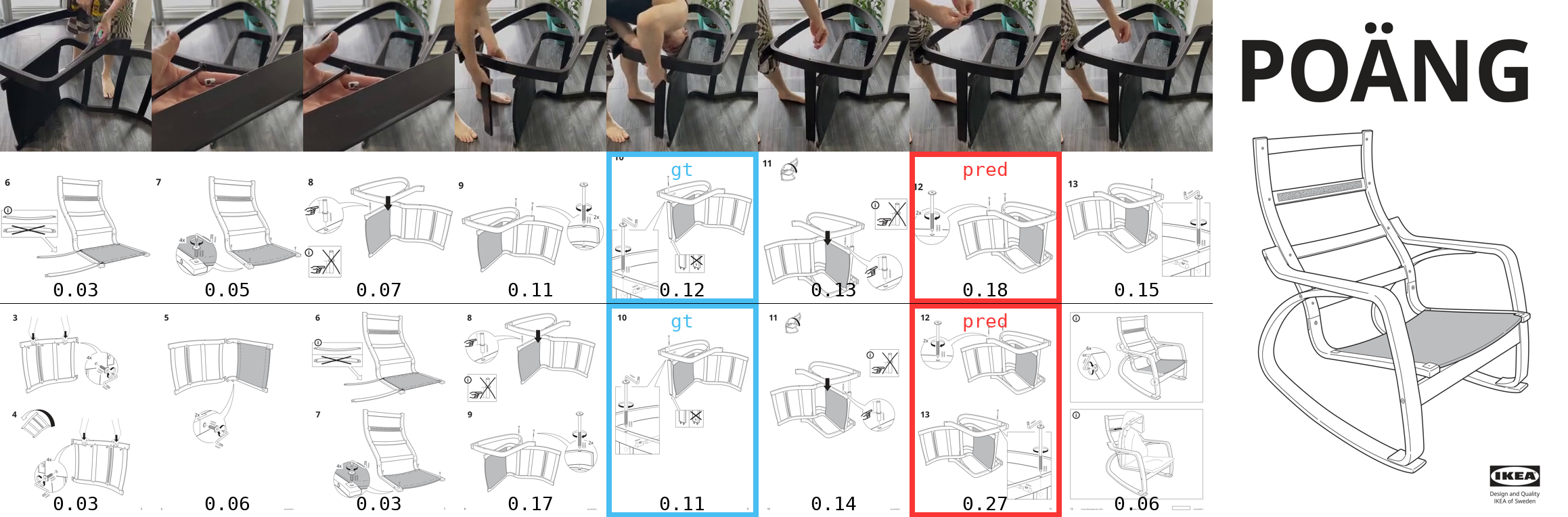}
        \caption{Alignment between a clip from YouTube video \href{https://www.youtube.com/watch?v=NXNxUE6XsF4}{NXNxUE6XsF4} and an Ikea furniture manual \href{https://www.ikea.com/au/en/p/poaeng-rocking-chair-white-stained-oak-veneer-skiftebo-dark-grey-s19395841/}{s19395841}.}
    \end{subfigure}
    \caption{Eight failure cases (4/8).}
    \label{fig:examples3}
\end{figure*}

\setcounter{figure}{6}

\begin{figure*}[ht]
    \centering
    \begin{subfigure}{0.86\linewidth}
        \addtocounter{subfigure}{4}
        \includegraphics[width=\linewidth]{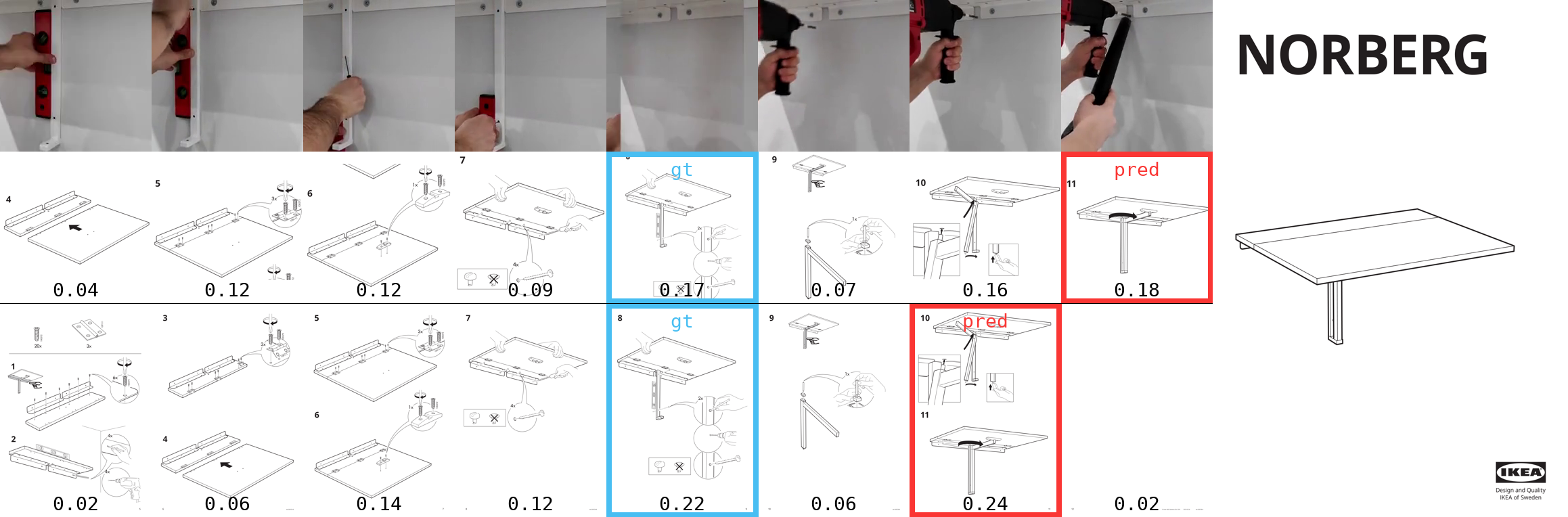}
        \caption{Alignment between a clip from YouTube video \href{https://www.youtube.com/watch?v=IUI8AQYTB0k}{IUI8AQYTB0k} and an Ikea furniture manual \href{https://www.ikea.com/au/en/p/norberg-wall-mounted-drop-leaf-table-white-90365793/}{90365793}.}
    \end{subfigure}
    \begin{subfigure}{0.86\linewidth}
        \includegraphics[width=\linewidth]{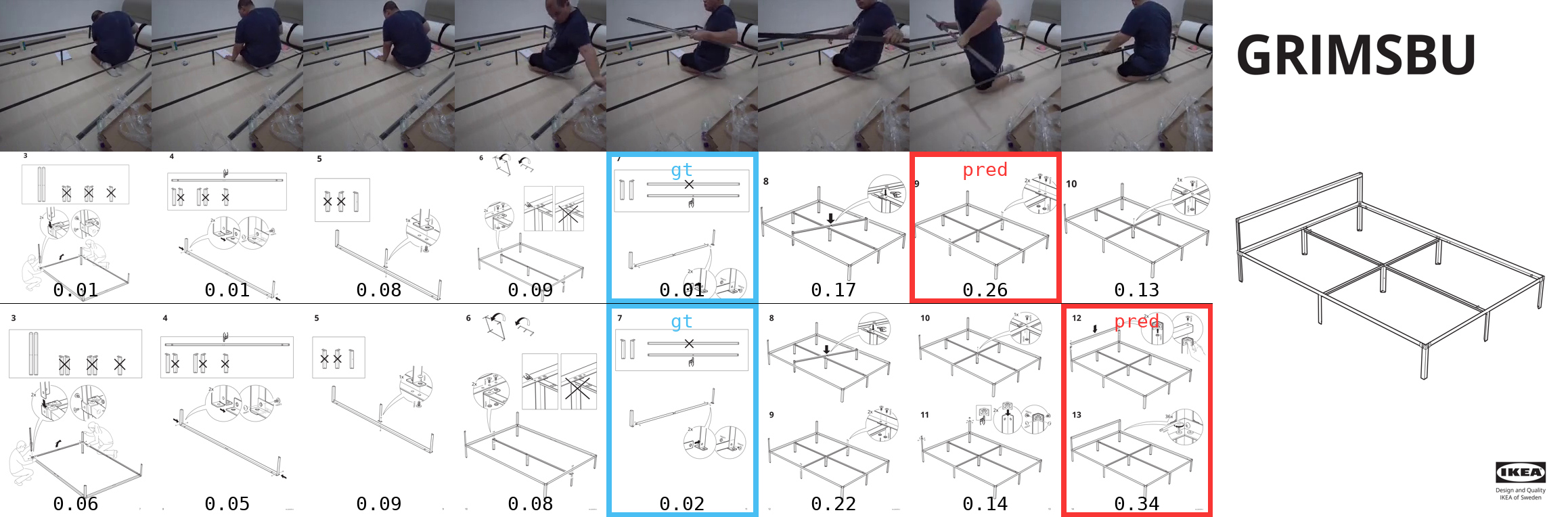}
        \caption{Alignment between a clip from YouTube video \href{https://www.youtube.com/watch?v=hFA07VckvOk}{hFA07VckvOk} and an Ikea furniture manual \href{https://www.ikea.com/au/en/p/grimsbu-bed-frame-grey-80458759/}{80458759}.}
    \end{subfigure}
    \begin{subfigure}{0.86\linewidth}
        \includegraphics[width=\linewidth]{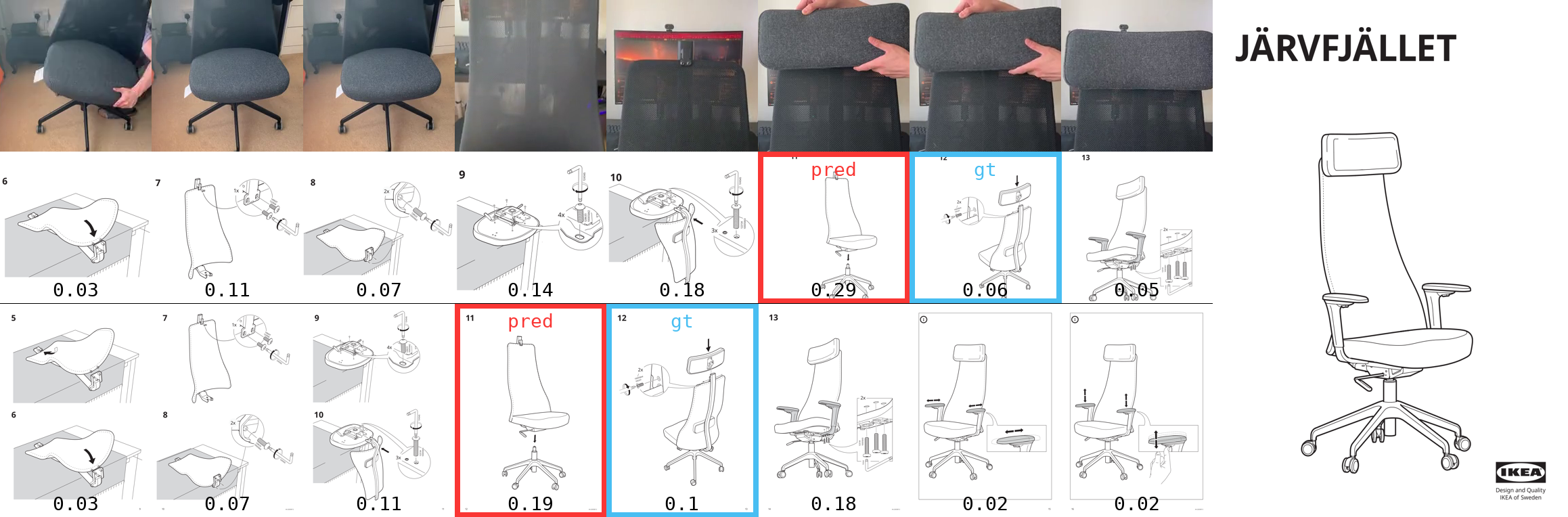}
        \caption{Alignment between a clip from YouTube video \href{https://www.youtube.com/watch?v=DSD9jawiz7o}{DSD9jawiz7o} and an Ikea furniture manual \href{https://www.ikea.com/au/en/p/jaervfjaellet-office-chair-with-armrests-glose-black-80494540/}{80494540}.}
    \end{subfigure}
    \begin{subfigure}{0.86\linewidth}
        \includegraphics[width=\linewidth]{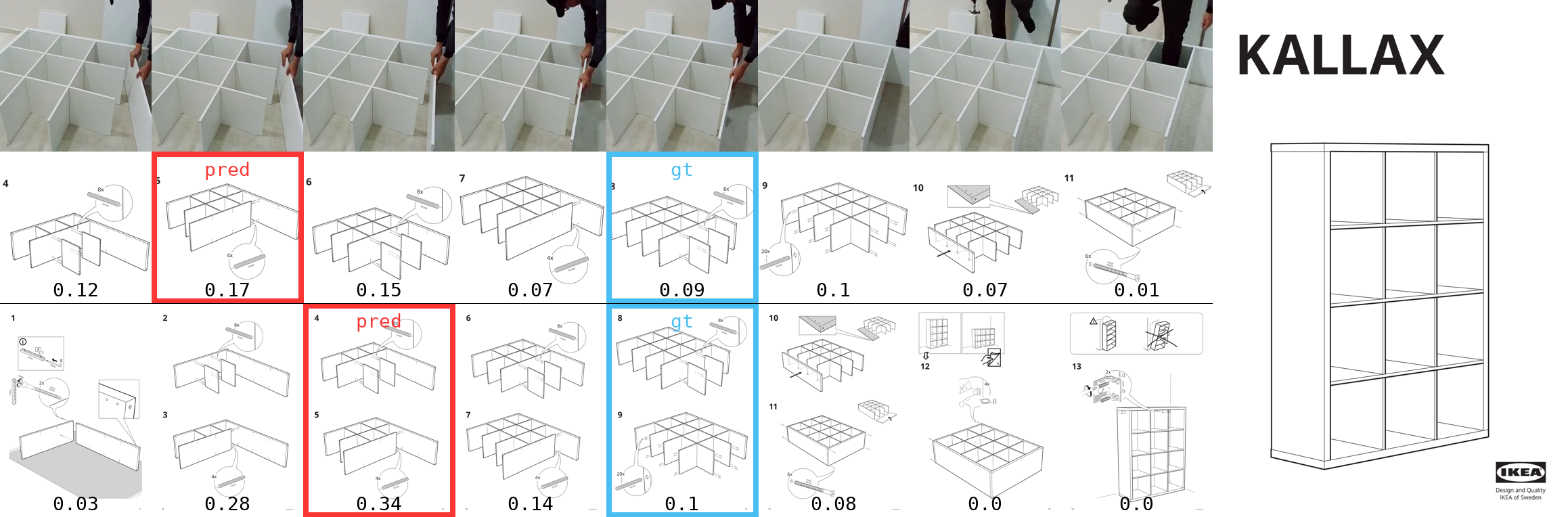}
        \caption{Alignment between a clip from YouTube video \href{https://www.youtube.com/watch?v=T6sQN6FtaDQ}{T6sQN6FtaDQ} and an Ikea furniture manual \href{https://www.ikea.com/au/en/p/kallax-shelving-unit-white-60409939/}{60409939}.}
    \end{subfigure}
    \caption{Eight failure cases (8/8).} 
    \label{fig:examples4}
\end{figure*}


{\small
\bibliographystyle{ieee_fullname}
\bibliography{supplementary}
}